\def\ARXIVPREPRINT{1}
\def\paperroot{.}
\documentclass{article}
\usepackage{iclr2027_conference}
\usepackage[T1]{fontenc}
\usepackage{times}
\usepackage{helvet}

\newif\ifarxivpreprint
\ifdefined\ARXIVPREPRINT
  \arxivpreprinttrue
  \iclrfinalcopy
\else
  \arxivpreprintfalse
\fi

\providecommand{\paperroot}{..}

\usepackage{amsmath,amsfonts,bm}

\def\eqref#1{equation~\ref{#1}}

\def\1{\bm{1}}

\DeclareMathAlphabet{\mathsfit}{\encodingdefault}{\sfdefault}{m}{sl}
\SetMathAlphabet{\mathsfit}{bold}{\encodingdefault}{\sfdefault}{bx}{n}

\usepackage{hyperref}
\hypersetup{hidelinks}
\usepackage{url}
\usepackage{booktabs}
\usepackage{graphicx}
\usepackage{float}
\input{result_layout.tex}
\usepackage{amsmath}
\usepackage{xcolor}
\usepackage{colortbl}  
\usepackage{tikz}
\usetikzlibrary{arrows.meta,positioning,fit,backgrounds,calc}
\definecolor{mDirect}{HTML}{1A6FB5}
\definecolor{mDirectL}{HTML}{DBE9F6}
\definecolor{mPred}{HTML}{D9700A}
\definecolor{mPredL}{HTML}{FBE9D6}

\definecolor{mVisE}{HTML}{0F7B8A}\definecolor{mVisF}{HTML}{DDF0F2} 
\definecolor{mActE}{HTML}{B85C00}\definecolor{mActF}{HTML}{FBE9D5} 
\definecolor{mStaE}{HTML}{2E7D46}\definecolor{mStaF}{HTML}{DFEFE4} 
\definecolor{mEffE}{HTML}{7A3FA6}\definecolor{mEffF}{HTML}{EEE3F7} 
\definecolor{mTrkE}{HTML}{1A6FB5}\definecolor{mTrkF}{HTML}{DBE9F6} 
\definecolor{mOutE}{HTML}{12507F}\definecolor{mOutF}{HTML}{C3DAEE} 
\definecolor{mCchE}{HTML}{4A5B8C}\definecolor{mCchF}{HTML}{E4E8F4} 
\definecolor{mHitE}{HTML}{C2185B}\definecolor{mHitF}{HTML}{FBE0EA} 
\definecolor{mAuxE}{HTML}{6B7280}\definecolor{mAuxF}{HTML}{F0F1F4} 
\definecolor{mLine}{HTML}{3C4450}
\definecolor{mTodo}{HTML}{C62828}   
\definecolor{uDrawback}{HTML}{9E3434} 
\definecolor{uNetworkF}{HTML}{1A6178} 
\usetikzlibrary{shapes.symbols}
\tikzset{udnetwork/.style={cloud,cloud puffs=12,cloud puff arc=120,
  aspect=2.5,draw=mLine,fill=uNetworkF,text=white,line width=.65pt,
  minimum width=54pt,minimum height=21pt,inner sep=1pt}}
\definecolor{mPanel}{HTML}{F7F9FC}                                 
\definecolor{mPanelE}{HTML}{E4EAF2}

\definecolor{uInputE}{HTML}{087E8B}\definecolor{uInputF}{HTML}{E4F3F3}
\definecolor{uModelE}{HTML}{42628A}\definecolor{uModelF}{HTML}{EAF0F8}
\definecolor{uOutputE}{HTML}{D55E00}\definecolor{uOutputF}{HTML}{FFF0E4}
\newcommand{\udeployfont}{\sffamily\fontsize{7}{8}\selectfont}
\newcommand{\udeployheadingfont}{\sffamily\fontsize{9}{10}\selectfont\bfseries}
\tikzset{
 udfont/.style={font=\udeployfont},
 udheading/.style={font=\udeployheadingfont},
 udbox/.style={mbox,udfont},
 udbaseline/.style={udbox,draw=mAuxE,fill=mPanel},
 udserver/.style={udbox,draw=uModelE,fill=uModelF},
 udedge/.style={udbox,draw=uOutputE,fill=uOutputF,line width=.8pt},
 udflow/.style={marr,draw=uOutputE,line width=.8pt},
 udremote/.style={marr,draw=uModelE,dashed,line width=.65pt}
}
\usetikzlibrary{decorations.pathreplacing}
\tikzset{udcostbrace/.style={draw=mLine,line width=.5pt,decorate,
 decoration={brace,mirror,amplitude=1.6pt}},
 udcostcomm/.style={fill=uInputE!35!white},
 udcostvla/.style={fill=uModelE},udcostlocal/.style={fill=uOutputE}}
\tikzset{uinput/.style={mbox,draw=uInputE,fill=uInputF},
 umodel/.style={mbox,draw=uModelE,fill=uModelF},
 uoutput/.style={mbox,draw=uOutputE,fill=uOutputF,line width=.8pt}}

\tikzset{
  mfont/.style      = {font=\sffamily\scriptsize},
  mbox/.style       = {draw=mLine, line width=0.5pt, rounded corners=2.2pt,
                       fill=white, inner sep=3.0pt, align=center,
                       font=\sffamily\scriptsize,
                       execute at begin node={\hyphenpenalty=10000\exhyphenpenalty=10000}},
  msmall/.style     = {mbox, font=\sffamily\tiny, inner sep=2.2pt},
  mvis/.style       = {mbox, draw=mVisE, fill=mVisF},
  mact/.style       = {mbox, draw=mActE, fill=mActF},
  msta/.style       = {mbox, draw=mStaE, fill=mStaF},
  meff/.style       = {mbox, draw=mEffE, fill=mEffF},
  mtrk/.style       = {mbox, draw=mTrkE, fill=mTrkF, line width=0.7pt},
  mout/.style       = {mbox, draw=mOutE, fill=mOutF, line width=0.7pt},
  mcch/.style       = {mbox, draw=mCchE, fill=mCchF},
  mhit/.style       = {mbox, draw=mHitE, fill=mHitF, line width=0.7pt},
  maux/.style       = {mbox, draw=mAuxE, fill=mAuxF, dashed,
                       dash pattern=on 1.5pt off 1.3pt, text=black!62},
  marr/.style       = {-{Stealth[length=1.6mm,width=1.4mm]}, draw=mLine,
                       line width=0.6pt},
  mthin/.style      = {-{Stealth[length=1.3mm,width=1.1mm]}, draw=mLine,
                       line width=0.4pt},
  mauxarr/.style    = {-{Stealth[length=1.3mm,width=1.1mm]}, draw=mAuxE,
                       line width=0.45pt, dashed, dash pattern=on 1.5pt off 1.3pt},
  mshare/.style     = {draw=mTrkE, line width=0.4pt, double, double distance=0.7pt},
  mhyph/.style      = {execute at begin node={\hyphenpenalty=10000\exhyphenpenalty=10000}},
  msub/.style       = {anchor=north west, font=\sffamily\scriptsize,
                       text=black!58, inner sep=0pt, mhyph},
  mtag/.style       = {anchor=west, font=\sffamily\tiny, text=black!58,
                       inner sep=1pt, mhyph},
  mtagc/.style      = {font=\sffamily\tiny, text=black!58, inner sep=1pt,
                       align=center, mhyph},
  mlab/.style       = {font=\sffamily\tiny\bfseries, inner sep=1pt},
  mcross/.style     = {mbox, draw=mAuxE, fill=mAuxF, text=black!45},
  mgroup/.style     = {draw=black!35, line width=0.4pt, rounded corners=2.4pt,
                       dashed, dash pattern=on 1.8pt off 1.5pt},
  taxroot/.style    = {mbox, draw=mLine, fill=mPanel, line width=0.8pt,
                       minimum width=4.2cm, minimum height=0.62cm,
                       font=\sffamily\scriptsize\bfseries},
  taxcall/.style    = {mbox, draw=mDirect, fill=mDirectL, line width=0.8pt,
                       minimum width=3.4cm, minimum height=0.58cm,
                       font=\sffamily\scriptsize\bfseries},
  taxinside/.style  = {mbox, draw=mAuxE, fill=mAuxF, line width=0.7pt,
                       minimum width=3.1cm, minimum height=0.58cm,
                       font=\sffamily\scriptsize\bfseries},
  taxleaf/.style    = {mbox, text width=2.75cm, minimum height=1.45cm,
                       inner sep=3.2pt},
  taxleafwide/.style= {mbox, text width=3.00cm, minimum height=1.45cm,
                       inner sep=3.2pt},
  taxours/.style    = {mbox, draw=mDirect, fill=mDirectL, line width=1.0pt,
                       minimum width=6.7cm, minimum height=0.58cm,
                       font=\sffamily\scriptsize\bfseries},
  taxlimit/.style   = {font=\sffamily\scriptsize\itshape, text=black!62,
                       align=center, inner sep=0pt},
  taxmethod/.style  = {font=\sffamily\scriptsize, text=black!68,
                       align=center, inner sep=0pt},
  taxline/.style    = {-{Stealth[length=1.5mm,width=1.3mm]}, draw=mLine,
                       line width=0.55pt},
  taxcompose/.style = {-{Stealth[length=1.4mm,width=1.2mm]}, draw=mAuxE,
                       line width=0.5pt, dashed,
                       dash pattern=on 1.7pt off 1.4pt},
}

\usetikzlibrary{svg.path}

\newcommand{\icoCamera}[2][mLine]{%
  \tikz[baseline={-0.5*#2pt-0.5ex},yscale=-1,sharp corners,scale={#2/24},line width={#2*2/24},line cap=round,line join=round,draw=#1]{%
    \draw svg {M5 7h1a2 2 0 0 0 2 -2a1 1 0 0 1 1 -1h6a1 1 0 0 1 1 1a2 2 0 0 0 2 2h1a2 2 0 0 1 2 2v9a2 2 0 0 1 -2 2h-14a2 2 0 0 1 -2 -2v-9a2 2 0 0 1 2 -2};
    \draw svg {M9 13a3 3 0 1 0 6 0a3 3 0 0 0 -6 0};
  }%
}

\def\micoh{8.0}                    

\tikzset{
  mico/.style = {baseline={-0.5*\micoh pt-0.5ex}, x=1pt, y=1pt, yscale=-1, sharp corners,
                 scale={\micoh/24}, line width={\micoh*2/24},
                 line cap=round, line join=round},
}

\newcommand{\mtrajico}[5]{\tikz[mico, draw=#1]{%
  \draw #2 -- #3 -- #4;
  \draw[-{Stealth[length=1.9,width=1.7]}] #4 -- #5;
  \foreach \p in {#2,#3,#4} {\fill[#1] \p circle (2.2);}}}
\newcommand{\icoTrajA}[1][mActE]{%
  \mtrajico{#1}{(2,19)}{(9,10)}{(15,13)}{(21.4,5.6)}}

\newcommand{\icoChunk}[1][mActE]{\icoTrajA[#1]}

\newcommand{\icoArm}[1][mStaE]{\tikz[mico, draw=#1]{%
  \draw (4.6,23.2) -- (13.4,23.2) -- (11.8,19.4) -- (6.2,19.4) -- cycle;
  \draw (9,19.4) -- (13.8,11.4) -- (20.6,15.2);
  \fill[#1] (9,19.4) circle (2.2);
  \fill[#1] (13.8,11.4) circle (2.2);
  \draw (20.6,15.2) circle (2.0);}}

\newcommand{\icoScene}[1][mLine]{\tikz[mico, draw=#1]{%
  \draw (1.5,21.5) -- (22.5,21.5);
  \draw[rounded corners={\micoh*1.4/24}] (14,14.5) rectangle (20,21.5);
  \draw (7,3.5) -- (7,10);
  \draw (3.5,10) -- (10.5,10);
  \draw (3.5,10) -- (3.5,15);
  \draw (10.5,10) -- (10.5,15);}}

\graphicspath{{figures/skip_success_2026-08-18/}}

\title{VLA-ULAP: Interleaving Cloud VLA Calls\\
with Ultra-Lightweight Local Action\\
Prediction at the Edge}

\ifarxivpreprint
\author{\textbf{Deyu Cao}$^{1}$ \quad \textbf{Ryuji Oi}$^{2}$ \quad
\textbf{Kosuke Matsushima}$^{2}$ \quad \textbf{Yuxuan Pan}$^{1}$ \\
\textbf{Ziheng Wang}$^{1}$ \quad \textbf{Daichi Fujiki}$^{2}$ \quad
\textbf{Atsutake Kosuge}$^{1}$ \\
{\normalfont $^{1}$The University of Tokyo \qquad
$^{2}$Institute of Science Tokyo}}
\hypersetup{
  pdfauthor={Deyu Cao, Ryuji Oi, Kosuke Matsushima, Yuxuan Pan, Ziheng Wang, Daichi Fujiki, Atsutake Kosuge},
  pdftitle={VLA-ULAP: Interleaving Cloud VLA Calls with Ultra-Lightweight Local Action Prediction at the Edge}
}

\else
  \author{Anonymous authors\\Paper under double-blind review}
\fi

\begin{document}
\maketitle
\ifarxivpreprint
  \lhead{Preprint}
\fi

\begin{abstract}
Billion-parameter vision--language--action (VLA) policies run either onboard,
consuming substantial power, or on remote servers, adding communication latency.
To address these drawbacks and better balance latency and onboard energy
consumption, we propose VLA-ULAP. It partitions inference across decision
times, interleaving remote VLA calls with predictions from an
Ultra-Lightweight Local Action Predictor (ULAP). A single ULAP has $\sim$7.4M parameters including
the frozen vision encoder. It combines current views, proprioception, and executed
action history to predict chunks in one pass. Trained independently, it
requires no VLA hidden states, online verification, or server round trips. On Jetson
Orin Nano, ULAP takes 20.7\,ms and 0.122\,J of idle-subtracted energy
per inference, versus 289.3\,ms and 40.46\,J for GR00T on RTX A6000.
Across four simulated base-policy/benchmark pairs, VLA-ULAP removes
45.3--77.3\% of VLA calls while retaining 95.0--98.5\% of baseline
success rates at selected operating points. On VLA-JEPA, VLA-ULAP also surpasses
local acceleration alternatives, using an estimated 49.4\% less inference time and 51.5\% less GPU
energy per successful episode than ACT, and 77.5\% less time and 80.7\% less
energy than SP-VLA at higher success rates.
In physical SO-101 trials, it similarly removes 70.3--71.9\% of VLA calls
without observed success-rate loss at seen or held-out placements.
Measured device costs imply 64.6--66.4\% less inference time and 70.1--71.7\%
less idle-subtracted energy per successful episode at these call counts.
Beyond these savings, faster responses help VLA-ULAP
exceed $\pi_{0.5}$'s success rate by 11.0 and 15.5 percentage points (pp)
on two tasks in latency-aware LIBERO-Safety simulation while approximately
halving VLA calls.
\end{abstract}

\section{Introduction}
\label{sec:intro}

Vision--language--action (VLA) policies leverage visual--language pretraining
and robot experience to follow instructions and generalize across tasks and environments
\citep{kim2024openvla,bjorck2025gr00tn1,black2025pi05}.
Deployment commonly uses onboard or remote GPUs.
Onboard inference avoids network delay, but trades power for responsiveness:
\citet{yang2026jetsonpi} report naive $\pi_{0.5}$ latencies of
76.4\,ms on RTX 4090 and 457.9\,ms on Jetson Thor, whose listed maximum
power ratings are 450\,W and 130\,W. The former's power budget challenges
battery-powered operation; the latter's latency spans about 14 control steps
at 30\,Hz, delaying responses to scene changes.
Remote inference offers powerful GPUs without their onboard battery burden,
but adds communication delay and jitter \citep{peng2026cloudedgevla}.
For example, \citet{huang2025daducorki} measure Wi-Fi image-transfer costs
of tens of ms and hundreds of mJ per frame.
Neither placement alone resolves this latency--onboard-energy trade-off,
raising the question:
\textbf{Can hybrid inference achieve a better balance between latency and
onboard energy consumption without sacrificing task success?}

\begin{figure}[t]
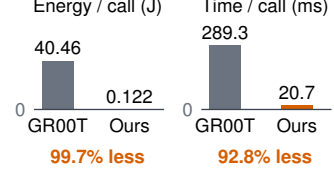

\centering
\input{figures/figure1_result_macros.tex}%
\providecommand{\paperroot}{..}
\providecommand{\ulapArtRoot}{figures/assets/ulap_deployment_generated_2026-09-15}%
\begin{tikzpicture}[udfont,x=1pt,y=1pt]
\providecommand{\udface}[3]{%
  \begin{scope}[shift={#1},draw=#2,line width=.6pt]
  \draw (0,0) circle (3.5pt);
  \fill[#2] (-1.25,1.3) circle (.35pt); \fill[#2] (1.25,1.3) circle (.35pt);
  \draw (-1.7,-1.3) .. controls (-.8,{-1.3-1.1*#3}) and (.8,{-1.3-1.1*#3}) .. (1.7,-1.3);
  \end{scope}%
}
\draw[mPanelE,line width=.65pt] (0,11) -- (391,11);
\node[udheading,anchor=west,fill=white,inner xsep=0pt,inner ysep=2pt]
  at (0,11) {Proposed and Conventional Inference Frameworks\quad};
\input{figures/fig_ulap_deployment_baselines.tex}
\input{figures/fig_ulap_deployment_hybrid.tex}
\input{figures/fig_ulap_deployment_benefits.tex}
\end{tikzpicture}
\caption{\textbf{Overview and benefits of the proposed VLA-ULAP framework.}
(a) Remote-only inference adds communication delay and energy cost, while
(b) full onboard VLA inference burdens the battery.
(c) VLA-ULAP schedules remote calls alongside local predictions.
Bottom panels show three benefits:
(1) more responsive control with higher dynamic-task success rates;
(2) lower estimated GPU-plus-edge energy and inference time per successful episode; and
(3) fast, low-energy inference on inexpensive edge devices for longer
battery life and scalable fleets.}
\label{fig:ulap-deployment}
\end{figure}

Existing methods accelerate individual policies but leave hybrid requirements
unmet. Quantization, token pruning,
and adaptive depth reduce work within a VLA call
\citep{zhang2026quantvla,liu2026vlapruner,yue2024deervla}, yet lower average
cost does not ensure that peak compute and memory demands fit a low-power
edge device. Adaptive execution horizons use more of an existing action
chunk \citep{feng2026dvac,liang2026aac}, but leave the VLA's per-call
computation unchanged and cannot extend execution beyond its generated
horizon. They therefore neither make the VLA fit a low-power edge device
nor remove the horizon-imposed limit on call reduction.
Lightweight policies such as ACT \citep{zhao2023act} generate actions locally, but without
large-scale pretraining, standalone deployment can fall short in task
success and generalization.
Model-internal partitioning instead couples a VLM with a fast action policy,
as in HiRT \citep{zhang2024hirt}. The fast policy uses VLM features,
so a remote VLM still requires communication to refresh them.

To achieve a better latency--onboard-energy trade-off through hybrid inference, we propose
\textbf{VLA-ULAP}, a hybrid framework that interleaves
occasional remote VLA calls with inexpensive action predictions from an
\textbf{Ultra-Lightweight Local Action Predictor (ULAP)} on the robot
(Figure~\ref{fig:ulap-deployment}). Local decisions respond to fresh observations
without a server round trip, while periodic VLA calls help preserve task
success by retaining access to the base policy. We partition inference across
decision times rather than within the model (e.g., a remote VLM and a local
action head), avoiding transfers of large intermediate features.
This framework offers three benefits:
(1) faster responses with fewer round trips,
(2) lower total inference time and energy, and
(3) affordable, low-power edge inference. ULAP avoids billion-parameter VLA
memory demands on edge devices, allowing inexpensive, low-power hardware
such as Jetson Orin Nano to replace consumer-grade GPUs. This supports
longer battery life and, together with fewer remote calls, larger robot
fleets per server.

To fully realize this framework's benefits, the local path must use fresh
observations, be lightweight, and remain independent of remote features or verification.
ULAP meets these requirements through the following design choices. It combines current camera views and
proprioception with the preceding executed action sequence, which
supplies motion context that appearance alone may not resolve.
A frozen Theia-Tiny encoder \citep{shang2024theia}, shallow multimodal fusion,
and an action head that generates the entire chunk in one pass without
iterative denoising keep prediction inexpensive and low-latency; the
model has 1.86--2.12M trainable parameters depending on the benchmark's
action chunk length, action dimensions, and task conditioning, plus 5.52M
frozen visual parameters (Figure~\ref{fig:method-ulap}).
Unlike HiRT~\citep{zhang2024hirt}, ULAP requires no intermediate features from the remote model.
The base VLA stays frozen. Only ULAP is
trained on demonstrations or successful base-policy trajectories; for a few
hundred episodes, this takes a few minutes on one A6000 GPU.
Because ULAP uses no VLA internal states, it requires no joint training
and does not depend on the base VLA's internal architecture.
At inference time, we interleave VLA calls and fresh local predictions at
chunk boundaries using fixed scheduling.

Our contributions are:
\begin{enumerate}
  \item A split inference framework that deploys local action prediction on a
  low-power edge device and reduces expensive VLA calls on remote servers. This improves the
  latency--onboard-energy trade-off while preserving task success.
  \item An ultra-lightweight local predictor that uses fresh vision, state, and
  action history to generate action chunks in one pass on a low-power edge device,
  without changing the base VLA or requiring its intermediate representations.
  \item Strong success retention across diverse robot-policy architectures,
  including VLA and world-model-based policies. Our evaluation covers manipulation
  on LIBERO~\citep{liu2023libero}, generalization to held-out objects in RoboCasa's
  varied kitchen environments~\citep{nasiriany2024robocasa}, and long-horizon,
  multi-task execution on CALVIN~\citep{mees2022calvin}.
  SO-101 trials~\citep{knight2024so101} and Jetson measurements additionally
  test real-robot performance and edge inference.
  Latency-aware LIBERO-Safety evaluation \citep{cui2026liberosafety} further
  tests whether fast local updates can improve success, not merely reduce cost.
\end{enumerate}
At selected simulation settings, 45.3--77.3\% fewer VLA calls retain
95.0--98.5\% of baseline success rate. In physical trials, mean VLA calls per
successful episode fall by 70.3--71.9\% without observed success-rate loss at ID or OOD placements.
On two dynamic tasks, our proposed VLA-ULAP improves success over
$\pi_{0.5}$ \citep{black2025pi05} by 11.0 and 15.5 percentage points (pp).
Section~\ref{sec:related} reviews related work, Section~\ref{sec:method}
describes our method, and Section~\ref{sec:howfar} presents the evaluation.

\section{Related Work}
\label{sec:related}

VLA inference acceleration methods can be broadly grouped into five families.

\paragraph{Reduce computation within a VLA call.}
Quantization \citep{zhang2026quantvla}, token pruning or reuse
\citep{liu2026vlapruner,xu2025vlacache}, and adaptive depth or intermediate-state
reuse \citep{yue2024deervla,liu2026latentbridge,oi2026actioncache} lower bit widths
or skip internal computation, yet low-latency inference on Jetson Orin
Nano-class devices can remain challenging.
They can complement VLA-ULAP's call reduction by reducing the computational
burden of the remaining VLA calls.

\paragraph{Adaptive execution chunk length.}
Action chunking amortizes one prediction over several control steps
\citep{zhao2023act}. AutoHorizon, AAC, and DVAC adapt the executed prefix
using attention, entropy, or denoising variance
\citep{wang2026autohorizon,liang2026aac,feng2026dvac}. Their selected actions
remain within a generated chunk, so call reduction is bounded by the base
policy's action chunk length.

\paragraph{Action reuse and extrapolation.}
FlashVLA repeats a recent action
\citep{tan2025flashvla}, whereas SP-VLA extrapolates an action-history buffer
\citep{li2025spvla}. They can act beyond the base policy's action chunk length,
but do not predict from current observations during bypass, so aggressive
reuse or extrapolation can miss scene changes and sharply degrade task success.
ULAP can also act beyond the VLA's action chunk length,
but predicts a new chunk conditioned on current observations and robot state.

\paragraph{Retrieve executable actions.}
RT-Cache \citep{kwon2025rtcache} uses frozen DINOv2 \citep{oquab2024dinov2}
and SigLIP \citep{zhai2023siglip} features to retrieve trajectory snippets.
ALT \citep{he2026alt} learns a query from multiple views and end-effector pose.
Retrieving actions can avoid expensive VLA inference, but their reported
zero-shot and OOD evaluations indicate limited generalization beyond memory
coverage, including changes in object placement.

\paragraph{Learn a cheaper action policy.}
HiRT conditions a fast policy on VLM-generated latents \citep{zhang2024hirt}.
Realtime-VLA FLASH couples an approximately 110M-parameter draft model to
the main image encoder and Action Expert verifier \citep{niu2026realtimeflash}.
ULAP requires neither transferred large-model features nor online verification.
We also compare an adapted ACT baseline \citep{zhao2023act} to test a standard
lightweight action-chunk predictor in the same hybrid framework.
Appendix~\ref{app:related-detail} expands the taxonomy.

\section{Method: VLA-ULAP}
\label{sec:method}

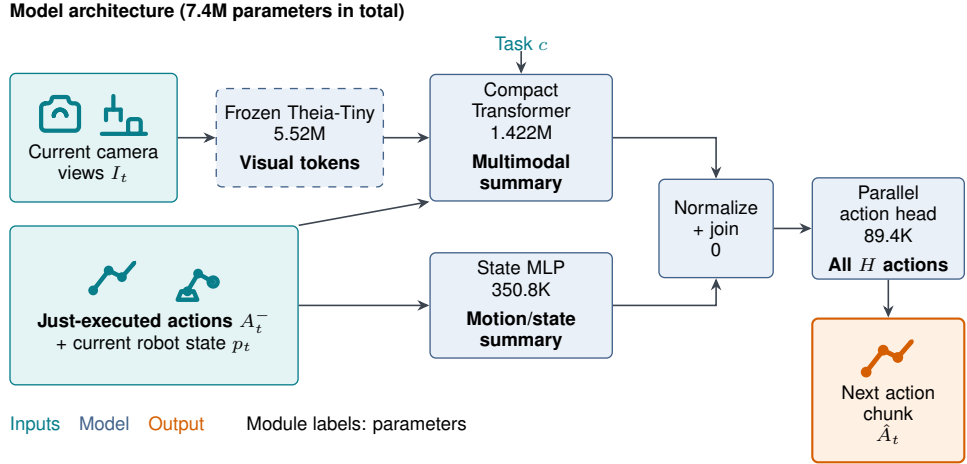
\begin{figure}[H]
\centering
\providecommand{\ulapOnlineSize}{7.4M}
\providecommand{\ulapVisionSize}{5.52M}
\providecommand{\ulapTrunkSize}{1.422M}
\providecommand{\ulapStateSize}{350.8K}
\providecommand{\ulapHeadSize}{89.4K}
\providecommand{\ulapFusionSize}{0}
\begin{tikzpicture}[node distance=5mm,
  mfont/.style={font={\udeployfont\baselineskip=8pt}},
  mbox/.append style={font={\udeployfont\baselineskip=8pt}},mfont]
\node[uinput,text width=2.0cm,minimum height=17mm] (images)
  {{\def\micoh{19}\icoCamera[uInputE]{19}\quad\icoScene[uInputE]}\\[2pt]
   Current camera\\views $I_t$};
\node[umodel,dashed,text width=2.0cm,minimum height=13mm,right=5mm of images] (theia)
  {Frozen Theia-Tiny\\\ulapVisionSize\\[3pt]\textbf{Visual tokens}};
\node[umodel,text width=2.2cm,minimum height=13mm,right=6mm of theia] (trunk)
  {Compact Transformer\\\ulapTrunkSize\\[3pt]\textbf{Multimodal summary}};
\node[mfont,text=uInputE,above=3mm of trunk,inner sep=0pt] (task) {Task $c$};
\node[mfont,anchor=south west,inner sep=0pt,yshift=2mm] (architecturelabel)
  at (images.west |- task.north)
  {\textbf{Model architecture (\ulapOnlineSize\ parameters in total)}};
\draw[marr] (task.south) -- (trunk.north);
\node[uinput,text width=3.6cm,minimum height=21mm,below=3mm of images.south west,anchor=north west] (history)
  {{\def\micoh{20}\icoChunk[uInputE]\qquad\icoArm[uInputE]}\\[2pt]
   \textbf{Just-executed actions} $A^-_t$\\+ current robot state $p_t$};
\node[umodel,text width=2.2cm,minimum height=13mm] (state)
  at (trunk.center |- history.center) {State MLP\\\ulapStateSize\\[3pt]\textbf{Motion/state summary}};
\node[umodel,text width=1.3cm,minimum height=13mm,right=6mm of trunk,yshift=-12mm] (fuse)
  {Normalize\\+ join\\\ulapFusionSize};
\node[umodel,text width=1.8cm,minimum height=13mm,right=5mm of fuse] (head)
  {Parallel action head\\\ulapHeadSize\\[3pt]\textbf{All $H$ actions}};
\node[uoutput,text width=1.8cm,minimum height=19mm,below=5mm of head] (output)
  {{\def\micoh{22}\icoChunk[uOutputE]}\\[2pt]Next action chunk\\$\hat A_t$};
\draw[marr] (images.east) -- (theia.west);
\draw[marr] (theia.east) -- (trunk.west);
\draw[marr] (history.east) -- (state.west);
\draw[marr] (history.north east) -- (trunk.south west);
\draw[marr] (trunk.east) -| (fuse.north);
\draw[marr] (state.east) -| (fuse.south);
\draw[marr] (fuse.east) -- (head.west);
\draw[marr] (head.south) -- (output.north);
\node[mfont,align=left,inner sep=0pt,below=4mm of history.south west,anchor=north west]
  (note) {\textcolor{uInputE}{Inputs}\quad\textcolor{uModelE}{Model}\quad\textcolor{uOutputE}{Output}
  \qquad Module labels: parameters};
\end{tikzpicture}
\caption{\textbf{ULAP model architecture.}
The Transformer summarizes vision, task, and motion context; the state branch
preserves nonvisual information. The action head uses their concatenated outputs
to predict a chunk in one pass. Counts are for five-step LIBERO-Safety.}
\label{fig:method-ulap}
\end{figure}

Building on the discussions in Section~\ref{sec:intro}, ULAP must use current observations,
remain lightweight, and operate without VLA hidden states or online verification.
The architecture in Figure~\ref{fig:method-ulap} implements these three
requirements as follows.

\paragraph{Predict chunks from local vision, state, and action history.}
At a decision boundary $t$, let $I_t$ denote the current camera views, $p_t$
the robot state, and $A^-_t=(a_{t-H},\ldots,a_{t-1})$ the preceding $H$
executed actions. ULAP predicts
\[
  \widehat A_t = g_\theta(I_t,p_t,A^-_t,c),
\]
where $c$ denotes task conditioning and $\widehat A_t$ contains the next $H$
actions. Similar views can occur while approaching, grasping, or withdrawing
from an object. To help distinguish these phases, ULAP combines current vision
and proprioception with the preceding actions actually sent to the robot,
regardless of whether the VLA or ULAP generated them.
The Transformer fuses $7\times7$-pooled Theia tokens
(Appendix~\ref{app:ulap-token-grid}), projected history/proprioception tokens,
and an aggregation token augmented with a learned task embedding.
A parallel state MLP maps
proprioception and flattened action history directly to a nonvisual summary.
Both summaries are independently normalized and concatenated with equal scaling.
The action head (Figure~\ref{fig:method-ulap}) applies a shared two-layer MLP
to fused features concatenated with learned temporal-position tokens,
predicting all $H$ actions in parallel.
GR00T input ablations support retaining both the state MLP and action history
for a better success--call-reduction trade-off at most tested
$\geq$90\%-retention settings (Appendix~\ref{app:ulap-input-ablations}).

\paragraph{Lightweight: 1.86--2.12M trainable parameters.}
ULAP combines a \textbf{5.52M-parameter frozen Theia-Tiny}
encoder~\citep{shang2024theia} with a \textbf{1.86--2.12M-parameter trainable
predictor}, depending on action chunk length, action dimensions, and task
conditioning. Including frozen vision, it is roughly 1/400 of the 3B-parameter
GR00T N1.7~\citep{nvidia2026gr00tn17}. Rather than running a large
vision--language backbone locally, ULAP uses compact visual features,
a shallow Transformer, and a small state MLP. Its MLP action head generates
the entire action chunk in a single forward pass without iterative denoising,
keeping prediction latency low.
Task embeddings or precomputed instruction features avoid an online language-model pass.
GR00T ablations find no significant success gain from Theia-Small, supporting
Tiny as the compact default (Appendix~\ref{app:ulap-theia-scale}).

\paragraph{Independent: direct prediction without VLA verification.}
All inputs to ULAP are available on the robot. Without VLA hidden states or
online verification, ULAP runs locally without feature-transfer or verification
round trips. The unchanged VLA is invoked only at scheduled decisions.

\paragraph{Training.}
Only the predictor is trained, while the VLA and Theia remain frozen.
We supervise the next action chunk with demonstrations or successful
base-policy rollouts using Smooth-L1 loss on training-set-standardized actions.

\paragraph{Scheduling and closed-loop execution.}
At deployment, the first decision invokes the VLA to initialize executed
history. We then use a fixed accumulator-based scheduler to spread local
decisions evenly among VLA calls. The requested local fraction $\rho$ is the
target proportion of chunk decisions assigned to ULAP rather than the VLA.
With equal execution horizons, the accumulator starts at $\rho$ to account for
the initial VLA chunk. At each subsequent boundary, it adds $\rho$;
if it reaches one, ULAP is selected and one is subtracted, otherwise the VLA
is called. This distributes VLA refreshes as regularly as the requested
fraction permits, avoiding unnecessary clusters of consecutive local predictions.
We also tested several adaptive scheduling signals, but found no clear,
consistent improvement over fixed scheduling and therefore adopt the fixed
scheduler as our default (Appendix~\ref{app:ulap-cads}).
Appendix~\ref{app:ulap-chunk-length} uses SmolVLA~\citep{shukor2025smolvla}
as a case study to show that VLA-ULAP retains higher success than simply executing
more of each VLA-generated chunk at comparable VLA-call reduction.

\paragraph{Lightweight ensemble.}
For accuracy, we equally average denormalized, clipped chunks from two ULAP
predictors (four for CALVIN), trained with different episode-level
training/validation partitions. Frozen Theia features are computed once and
shared; only lightweight predictor passes and averaging add computation.
The VLA and fixed schedule remain unchanged.

\raggedbottom
\section{Results}
\label{sec:howfar}

We test two linked benefits: retaining success with less VLA computation,
and improving success when delayed actions encounter a changing scene.
Section~\ref{sec:ulap-simulation} tests four base-policy/benchmark pairs,
including long-horizon task chains.
Section~\ref{sec:ulap-spvla} compares
success, inference speedup, and energy efficiency against ACT-based local
prediction, SP-VLA action extrapolation, and standalone RT-Cache retrieval.
Section~\ref{sec:ulap-real-robot} tests whether VLA-ULAP remains effective on a real robot,
while Section~\ref{sec:ulap-jetson} separately profiles inference on Jetson Orin Nano
and A6000. Section~\ref{sec:ulap-safety} returns to simulation to test whether
lower latency also improves success in dynamic environments.

\subsection{Simulation across base policies and benchmarks}
\label{sec:ulap-simulation}
\label{sec:ulap-libero-groot}
\label{sec:ulap-libero-jepa}
\label{sec:ulap-robocasa-cosmos}

\paragraph{LIBERO $\times$ GR00T N1.7.}
We first pair GR00T N1.7~\citep{nvidia2026gr00tn17} with ULAP using fixed scheduling and
16-step chunks on all 40 tasks in LIBERO, a simulated robot-manipulation
benchmark~\citep{liu2023libero}.
Training and evaluation initial states are disjoint
(Appendix~\ref{app:groot-ulap-fixed}).
In Figure~\ref{fig:ulap-fixed}, single-predictor VLA-ULAP achieves \textbf{88.50\%} success
(\textbf{95.0\%} retention) with \textbf{50.7\%} fewer VLA calls.
The ensemble achieves \textbf{88.50\%} success (\textbf{95.0\%} retention)
with \textbf{50.9\%} fewer calls.
Ensembling adds up to \textbf{2.75~pp} at high local fractions,
with a mean gain of \textbf{0.99~pp} across all nine tested fractions.

\paragraph{LIBERO $\times$ VLA-JEPA.}
To test another architecture, we pair VLA-JEPA~\citep{sun2026vlajepa}
with ULAP using fixed scheduling and seven-step chunks on
all 40 tasks, using disjoint training and evaluation initial states as in GR00T
(Appendix~\ref{app:jepa-ulap-fixed}).
In Figure~\ref{fig:jepa-ulap-fixed}, the single predictor and ensemble achieve
\textbf{95.35\%} and \textbf{96.30\%} success (\textbf{96.5\%} and \textbf{97.5\%}
retention) with \textbf{77.0\%} and \textbf{77.3\%} fewer VLA calls, respectively.
Ensembling adds up to \textbf{2.35~pp} at high
local fractions, with a mean gain of \textbf{0.68~pp} across all nine settings.

\begin{figure}[t]
\centering
\resultpanel{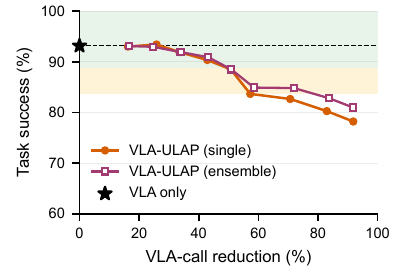}
{\textbf{LIBERO $\times$ GR00T N1.7.} Call reduction is relative to VLA only (star/dashed line).
Green/amber bands show $\geq$95\%/90--95\% success-rate retention.}
{fig:ulap-fixed}\hfill
\resultpanel{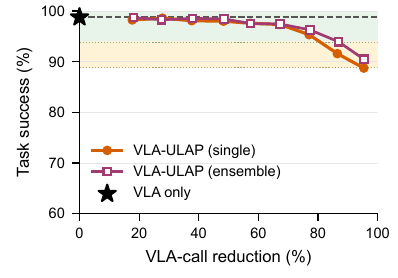}
{\textbf{LIBERO $\times$ VLA-JEPA.}
Markers and bands follow Figure~\ref{fig:ulap-fixed}.}
{fig:jepa-ulap-fixed}
\end{figure}

\resultparagraph{RoboCasa $\times$ Cosmos-Policy}{}
{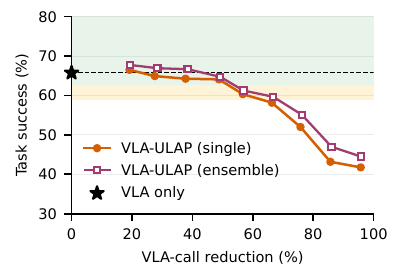}
{\textbf{RoboCasa $\times$ Cosmos-Policy.}
Bands follow Figure~\ref{fig:ulap-fixed}.}
{fig:cosmos-ulap-fixed}
{We next test more varied kitchen layouts and object placements in
RoboCasa~\citep{nasiriany2024robocasa}. With Cosmos-Policy~\citep{kim2026cosmospolicy},
we train a shared ULAP on 24 tasks using object split A and evaluate on
split B, testing generalization to held-out objects
(Appendix~\ref{app:cosmos-ulap-fixed}).

Single-predictor and ensemble VLA-ULAP retain \textbf{97.5\%} and \textbf{98.5\%} of the baseline
success rate (\textbf{64.08\%} and \textbf{64.75\%}), removing
\textbf{48.8\%} and \textbf{49.2\%} of calls, respectively
(Figure~\ref{fig:cosmos-ulap-fixed}). Ensemble gains span \textbf{0.67--3.83~pp}
across nine local fractions.}

\resultparagraph{CALVIN $\times$ X-VLA}{}
{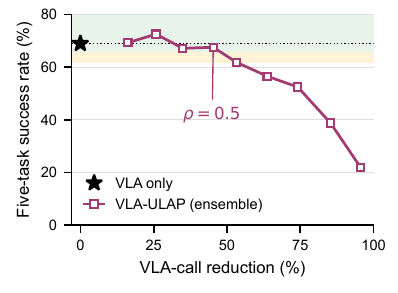}
{\textbf{CALVIN $\times$ X-VLA.} Five-task success rate. Bands follow Figure~\ref{fig:ulap-fixed}.}
{fig:calvin-ulap-tradeoff}
{Additionally, we use CALVIN~\citep{mees2022calvin} to test long-horizon,
multi-task execution through five instructions without resets.
With X-VLA~\citep{zheng2026xvla} and a four-model ULAP
ensemble at $\rho=0.5$, VLA-ULAP removes \textbf{45.3\%} of calls while
retaining \textbf{98.0\%} of baseline success rate (67.6\% versus 69.0\%;
Figure~\ref{fig:calvin-ulap-tradeoff}).
Mean completed tasks are maintained (4.176 versus 4.138), supporting
effective bypass across long task chains. Appendix~\ref{app:calvin-ulap}
provides depth-wise results and settings.}

\subsection{Comparison with prior work}
\label{sec:ulap-spvla}

Having tested VLA-ULAP across models and benchmarks, we compare three alternatives.
\textbf{ACT}~\citep{zhao2023act}, a vision-based action-chunk imitation policy,
tests a standard learned local path. \textbf{SP-VLA}~\citep{li2025spvla}
tests action-history extrapolation using its scheduling component without
token pruning. Both can supply actions beyond the VLA's last generated chunk;
VLA-ULAP, ACT, and SP-VLA share the same VLA-JEPA base policy.
\textbf{RT-Cache}~\citep{kwon2025rtcache} instead tests standalone image-based
action retrieval, using the same training rollout corpus without VLA calls.
All methods are evaluated on the same LIBERO initial states.

\begin{figure}[H]
\centering
\includegraphics{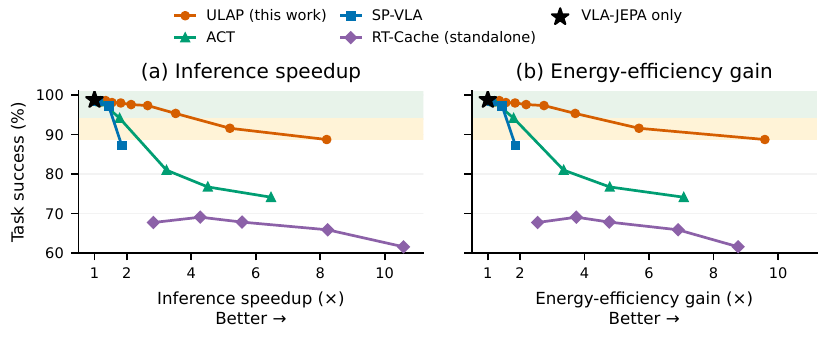}
\caption{\textbf{Comparison with prior methods on LIBERO: success and inference efficiency.} Gains are VLA-JEPA-only cost divided by method
cost, averaging successful episodes and using measured RTX A6000 branch costs.
Larger gains lie to the right; VLA-JEPA alone is $1\times$.
RT-Cache runs standalone with chunk lengths 4/6/8/12/16.
Bands follow Figure~\ref{fig:ulap-fixed}; settings are in
Appendix~\ref{app:jepa-spvla}.}
\label{fig:jepa-spvla}
\end{figure}

Figure~\ref{fig:jepa-spvla} compares success with inference speedup and
energy-efficiency gain over VLA-JEPA, using mean costs per successful episode.
VLA-ULAP achieves a better observed frontier than all three alternatives---ACT,
SP-VLA, and RT-Cache.
Compared with ACT at 94.15\% success, VLA-ULAP achieves \textbf{95.35\%}
with \textbf{49.4\% less inference time and 51.5\% less GPU energy}.
Outperforming ACT's 51.6M-parameter model supports ULAP's compact design.
Compared with the most aggressive SP-VLA setting at 87.25\% success,
VLA-ULAP achieves 88.75\% with \textbf{77.5\% less time and 80.7\% less energy}.
SP-VLA's sharp decline with longer observation-free extrapolation
highlights the value of using fresh observations.
Compared with standalone RT-Cache's peak success of \textbf{69.10\%}
(chunk length 6), VLA-ULAP achieves \textbf{91.60\%} with
\textbf{17.7\% less time and 34.2\% less energy}.
RT-Cache's low success on held-out
placements suggests limited generalization beyond cache coverage.

\subsection{Real robot $\times$ GR00T N1.7}
\label{sec:ulap-real-robot}

Moving from simulation to a real robot, we evaluate two pick-and-place tasks
on an SO-101 arm~\citep{knight2024so101}. The robot places a ping-pong ball
into a bowl in one task and a glue stick into a cup in the other.
Each task uses 100 human demonstration episodes to fine-tune GR00T;
ULAP uses 80 for training, 10 for validation, and 10 for testing.
Given the ensemble gains in simulation, we use
ensemble ULAP with modest latency and energy overhead
(Appendix~\ref{app:real-robot-ulap}, Table~\ref{tab:jetson-single-ensemble}).
Local ULAP and cloud VLA both execute 16-step chunks, with requested local fraction 0.75.
Each policy is evaluated on 75 trials, comprising 50 at training placements
(ID) and 25 at held-out placements (OOD).

\textbf{Performance is retained at both in-distribution and held-out placements.}
VLA-ULAP matches GR00T-only's OOD success rates of \textbf{88.0\%} (22/25)
on Ping-pong and \textbf{100.0\%} (25/25) on Glue-stick, while achieving
\textbf{96.0\%} (48/50) at ID placements on both tasks.

\begin{figure}[H]
\centering
\includegraphics[trim=0 12bp 0 0,clip]{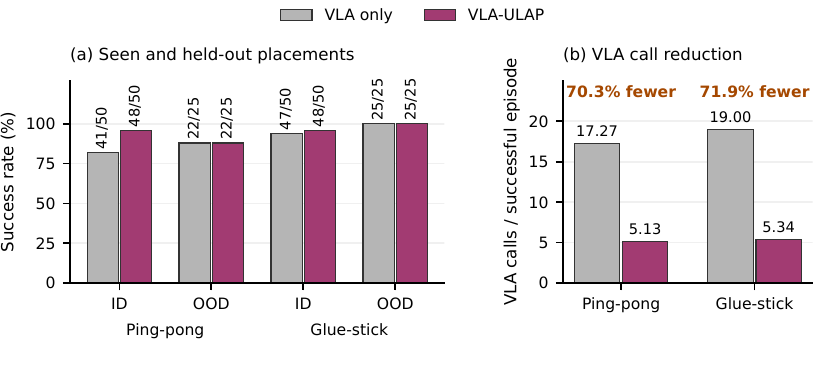}
\caption{\textbf{Real-robot evaluation on SO-101 with ensemble ULAP.} Left: ID/OOD success rates, with successful/total trial counts.
Right: mean VLA calls per successful episode and reduction from GR00T-only.
Settings: Appendix~\ref{app:real-robot-ulap}.}
\label{fig:real-robot-ulap}
\end{figure}

\resultbeside{%
\begin{minipage}[t]{.49\linewidth}
\centering
\includegraphics[width=\linewidth]{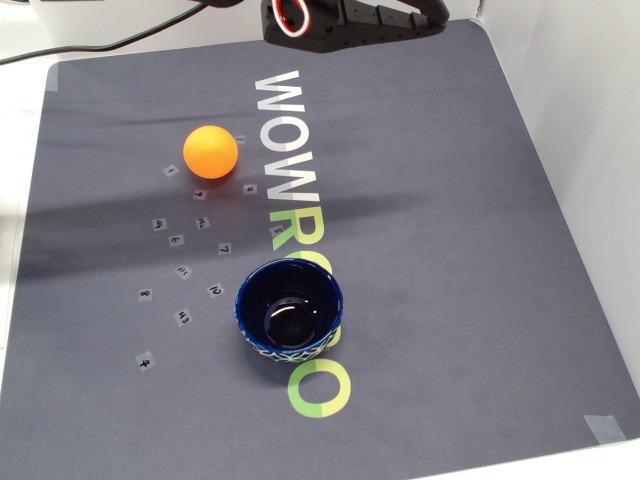}

\small Ping-pong: ball $\rightarrow$ blue bowl
\end{minipage}\hfill
\begin{minipage}[t]{.49\linewidth}
\centering
\includegraphics[width=\linewidth]{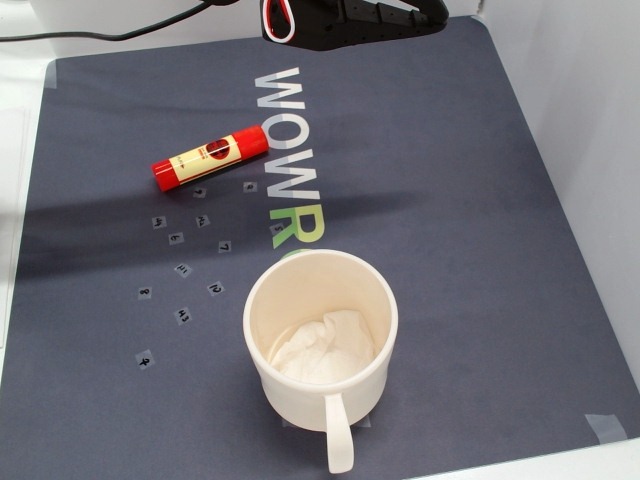}

\small Glue-stick: glue stick $\rightarrow$ white cup
\end{minipage}
\caption{\textbf{Physical pick-and-place tasks.} Initial external-camera views
from evaluated trials.}
\label{fig:real-robot-ulap-tasks}
}{%
Mean VLA calls per successful episode fall by \textbf{70.3\%} on Ping-pong
and \textbf{71.9\%} on Glue-stick (Figure~\ref{fig:real-robot-ulap}),
with no observed success-rate decrease at either ID or OOD placements.
}

\subsection{A lightweight local fast path}
\label{sec:ulap-jetson}

\begin{figure}[!htbp]
\centering
\includegraphics{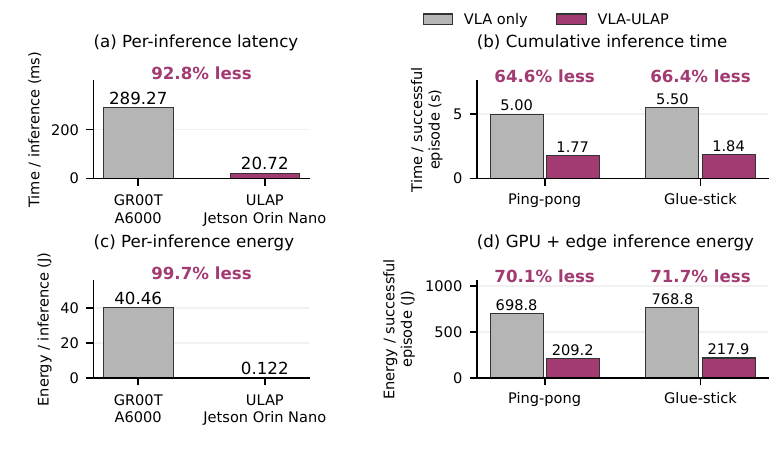}
\caption{\textbf{Inference latency and energy for real-robot evaluation.}
Left: measured costs of GR00T on A6000 and ensemble ULAP on Jetson Orin Nano.
Right: estimated cumulative inference time and energy per successful episode.
Energy sums idle-subtracted A6000 board and Jetson module costs;
details are in Appendix~\ref{app:ulap-jetson-cost}.}
\label{fig:ulap-jetson-cost}
\end{figure}

Physical rollouts used Mac MPS for ULAP; separate profiling measures its
edge-device cost. On Jetson Orin Nano, ensemble ULAP takes \textbf{20.72\,ms}
and \textbf{0.122\,J} of idle-subtracted energy per inference, versus
\textbf{289.27\,ms} and \textbf{40.46\,J} for GR00T on A6000.
This saves \textbf{92.8\%} of inference time and \textbf{99.7\%} of inference energy per local decision.
Combining these measured costs with successful physical episodes' call counts,
we estimate inference time reductions of \textbf{64.6\%} on Ping-pong and
\textbf{66.4\%} on Glue-stick, with idle-subtracted GPU-plus-edge energy reductions of \textbf{70.1\%}
and \textbf{71.7\%}, respectively (Figure~\ref{fig:ulap-jetson-cost}).
Thus, even on an inexpensive edge device, the local path delivers fast
inference while substantially reducing inference energy consumption.

\subsection{LIBERO-Safety $\times$ $\pi_{0.5}$: beyond faster execution}
\label{sec:ulap-safety}

Finally, we ask whether the fast path offers more than computational savings:
when the scene changes during inference, faster local updates may also improve
task success.
We test this benefit on two latency-sensitive tasks from LIBERO-Safety's
dynamic obstacle-avoidance suite \citep{cui2026liberosafety}: placing both moka
pots on a stove, and placing a mug in a microwave and closing its door.
We model three control-step delays for the VLA and zero additional steps
for local prediction, reflecting their measured latency scales
(Appendix~\ref{app:ulap-safety}). The simulator continues stepping during
the modeled VLA delay.
We compare $\pi_{0.5}$-only execution \citep{black2025pi05} with VLA-ULAP replacing half of the
scheduled decisions, using the same five-step execution interval and
200 episodes per task and policy (Appendix~\ref{app:ulap-safety}).

\begin{table}[H]
\centering
\caption{\textbf{Evaluation on two LIBERO-Safety tasks under latency-aware execution.}
Each condition uses 50 initial states with four repetitions.
Gain is the absolute success-rate increase in pp.}
\label{tab:ulap-safety}
\begin{tabular}{lrrr}
\toprule
Task & $\pi_{0.5}$ only & VLA-ULAP & Gain \\
\midrule
Moka pots $\rightarrow$ stove & 77.5\% (155/200) & \textbf{88.5\%} (177/200) & \textbf{+11.0 pp} \\
Mug $\rightarrow$ microwave & 50.0\% (100/200) & \textbf{65.5\%} (131/200) & \textbf{+15.5 pp} \\
\bottomrule
\end{tabular}

\end{table}

VLA-ULAP raises success by \textbf{11.0} and \textbf{15.5~pp},
while reducing total VLA calls by \textbf{51.8\%} and \textbf{52.5\%},
respectively (Table~\ref{tab:ulap-safety}). Thus, the fast local path does not
merely trade accuracy for cheaper inference: on these latency-sensitive
tasks, it achieves both fewer heavy-policy calls and more successful executions.

\section{Conclusion}
\label{sec:conclusion}
VLA-ULAP combines remote VLA calls with ultra-lightweight local prediction to
reduce communication delay and onboard energy demand. ULAP generates fresh
chunks from current observations, proprioception, and executed action history,
while periodic VLA calls help preserve success. Independence from VLA hidden
states and online verification enables low-power edge deployment without
joint training or costly transfers of large intermediate features during inference.

Across four simulation settings, selected points remove 45.3--77.3\% of VLA
calls while retaining 95.0--98.5\% of baseline success rate. On VLA-JEPA,
VLA-ULAP achieves a stronger success--efficiency trade-off than
three representative approaches to reducing VLA calls. Physical SO-101
tests reduce VLA calls without observed success-rate loss under placement shifts,
and Jetson Orin Nano ensemble measurements
show the feasibility of fast, low-energy local inference. Beyond these savings, latency-aware
LIBERO-Safety results show that faster updates can improve dynamic-task success.

Together, these results support decision-time partitioning that retains VLA capability
while reducing remote requests. Low-power, low-latency local prediction enables
responsive control and longer battery life, while freeing shared-GPU capacity
for concurrent operation of larger fleets.
\label{main-text-end}

\clearpage
\section*{AI Use Statement}
Generative AI tools assisted with developing the conceptual framing,
refining hypotheses, providing feedback on method and experimental design,
interpreting reported results, and translating author notes into English
manuscript text. They also assisted with
implementing ULAP and its training and evaluation code, as well as
organizing and reformatting evaluation logs and implementing scripts for
data auditing, aggregation, cost calculations, and figure generation.
AI agents were also used to execute experimental commands, manage experiments,
and diagnose and fix errors.
Additional uses included literature discovery and synthesis, manuscript
drafting, restructuring and editing, reference formatting, and creation and
revision of scientific figures and conceptual illustrations.
No generative-AI-generated synthetic data were used for training or evaluation.
Generative AI was not used to formulate mathematical or theoretical claims
or to write or verify proofs.

The authors reviewed the AI-assisted text, citations, result interpretations,
and figures, and rigorously checked and tested the implementation code,
including training and evaluation code.
Numerical reporting is checked against recorded evaluation artifacts,
and cited claims are checked against primary sources documented in a source
ledger. Analysis scripts include consistency checks on aggregate counts and
derived metrics, and rendered figures and manuscript layouts are inspected.
The authors retain responsibility for the final text, methods, results,
code, figures, and all AI-assisted claims and artifacts.

\section*{Ethics Statement}
Learned robot policies can produce unsafe actions when observations are
unfamiliar or inaccurate. Our physical evaluation concerns bounded tabletop
pick-and-place, retaining the existing joint-command clamps and episode timeout
(Appendix~\ref{app:real-robot-ulap}). Neither success at held-out placements nor
the LIBERO-Safety simulation establishes safe human--robot collaboration.
ULAP is not a safety controller;
deployment near people requires independent safeguards and risk assessment.
Reported energy savings concern inference devices, not the robot's full
operational or life-cycle footprint.

\section*{Reproducibility Statement}
Section~\ref{sec:method} specifies the predictor inputs, architecture, training
objective, and scheduling. Appendices~\ref{app:ulap-fixed} and
\ref{app:ulap-cads} give benchmark-specific training and evaluation settings,
seeds, action horizons, and metric definitions. Appendix~\ref{app:ulap-safety}
describes injected-delay execution; Appendix~\ref{app:ulap-jetson-cost}
states the hardware measurement boundaries and episode-cost reconstruction.
Per-task plots and success counts accompany these settings so that aggregate
claims can be checked without conflating protocols.

\bibliography{refs}
\bibliographystyle{iclr2027_conference}

\clearpage
\appendix
\section*{Appendix}
\suppressfloats[t]
\section{Extended Related-Work Taxonomy}
\label{app:related-detail}
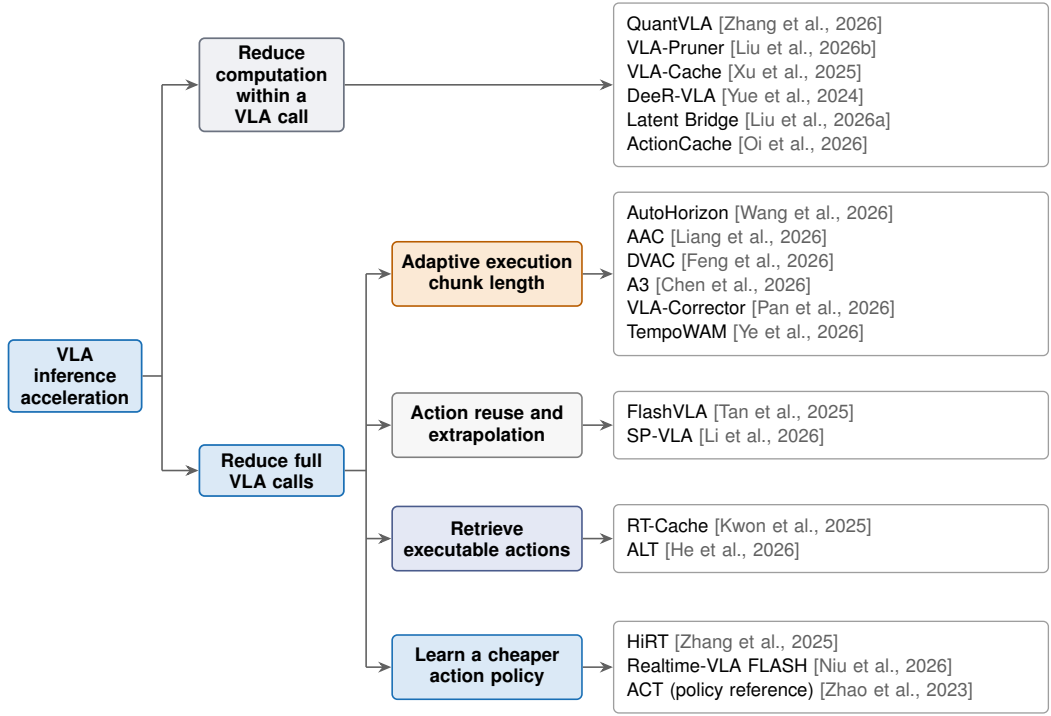
\begin{figure}[t]
\centering
\providecommand{\taxcite}[2]{\href{#1}{\textcolor{black!62}{[#2]}}}
\begin{tikzpicture}[
  x=1cm,y=1cm,
  every node/.style={font=\sffamily\scriptsize,align=center},
  axis/.style={rounded corners=2pt,draw=mDirect,fill=mDirectL,
    line width=.65pt,text width=1.7cm,inner sep=3pt,font=\sffamily\scriptsize\bfseries},
  family/.style={rounded corners=2pt,draw=black!50,fill=black!3,
    line width=.6pt,text width=2.3cm,minimum height=.85cm,
    inner sep=3pt,font=\sffamily\scriptsize\bfseries},
  papers/.style={rounded corners=2pt,draw=black!38,fill=white,
    line width=.5pt,text width=5.4cm,inner sep=5pt,align=left},
  arr/.style={-{Stealth[length=1.75mm,width=1.45mm]},draw=black!62,line width=.6pt}
]
\node[axis,text width=1.55cm] (root) at (.75,.65) {VLA\\inference\\acceleration};
\node[axis,draw=mAuxE,fill=mAuxF] (incall) at (3.35,4.5)
  {Reduce\\computation\\within a VLA call};
\node[axis] (calls) at (3.35,-.6) {Reduce full\\VLA calls};
\node[family,draw=mActE,fill=mActF] (horizon) at (6.2,2)
  {Adaptive execution\\chunk length};
\node[family] (reuse) at (6.2,0) {Action reuse and\\extrapolation};
\node[family,draw=mCchE,fill=mCchF] (retrieve) at (6.2,-1.5)
  {Retrieve\\executable actions};
\node[family,draw=mDirect,fill=mDirectL] (learn) at (6.2,-3.2)
  {Learn a cheaper\\action policy};
\coordinate (rootbranch) at (1.9,.65);
\draw[draw=black!62,line width=.6pt] (root.east) -- (rootbranch);
\draw[draw=black!62,line width=.6pt]
  (rootbranch |- incall.west) -- (rootbranch |- calls.west);
\foreach \objective in {incall,calls}
  \draw[arr] (rootbranch |- \objective.west) -- (\objective.west);
\coordinate (branch) at (4.6,-.6);
\draw[draw=black!62,line width=.6pt] (calls.east) -- (branch);
\draw[draw=black!62,line width=.6pt] (4.6,2) -- (4.6,-3.2);
\foreach \family in {horizon,reuse,retrieve,learn}
  \draw[arr] (branch |- \family.west) -- (\family.west);

\node[papers] (insidepapers) at (10.75,4.5) {
  QuantVLA \taxcite{https://openaccess.thecvf.com/content/CVPR2026/html/Zhang_QuantVLA_Scale-Calibrated_Post-Training_Quantization_for_Vision-Language-Action_Models_CVPR_2026_paper.html}{Zhang et al., 2026}\\[1pt]
  VLA-Pruner \taxcite{https://arxiv.org/abs/2511.16449}{Liu et al., 2026b}\\[1pt]
  VLA-Cache \taxcite{https://proceedings.neurips.cc/paper_files/paper/2025/hash/f062da1973ac9ac61fc6d44dd7fa309f-Abstract-Conference.html}{Xu et al., 2025}\\[1pt]
  DeeR-VLA \taxcite{https://proceedings.neurips.cc/paper_files/paper/2024/hash/67b0e7c7c2a5780aeefe3b79caac106e-Abstract-Conference.html}{Yue et al., 2024}\\[1pt]
  Latent Bridge \taxcite{https://arxiv.org/abs/2605.02739}{Liu et al., 2026a}\\[1pt]
  ActionCache \taxcite{https://arxiv.org/abs/2607.06370}{Oi et al., 2026}
};
\node[papers] (horizonpapers) at (10.75,2) {
  AutoHorizon \taxcite{https://arxiv.org/abs/2602.21445}{Wang et al., 2026}\\[1pt]
  AAC \taxcite{https://openaccess.thecvf.com/content/CVPR2026/html/Liang_Adaptive_Action_Chunking_at_Inference-time_for_Vision-Language-Action_Models_CVPR_2026_paper.html}{Liang et al., 2026}\\[1pt]
  DVAC \taxcite{https://arxiv.org/abs/2606.03847}{Feng et al., 2026}\\[1pt]
  A3 \taxcite{https://arxiv.org/abs/2605.11567}{Chen et al., 2026}\\[1pt]
  VLA-Corrector \taxcite{https://arxiv.org/abs/2607.01804}{Pan et al., 2026}\\[1pt]
  TempoWAM \taxcite{https://arxiv.org/abs/2608.09492}{Ye et al., 2026}
};
\node[papers] (reusepapers) at (10.75,0) {
  FlashVLA \taxcite{https://arxiv.org/abs/2505.21200}{Tan et al., 2025}\\[1pt]
  SP-VLA \taxcite{https://proceedings.iclr.cc/paper_files/paper/2026/hash/4072543747a14bbed76284cf2c04b9e9-Abstract-Conference.html}{Li et al., 2026}
};
\node[papers] (retrievepapers) at (10.75,-1.5) {
  RT-Cache \taxcite{https://doi.org/10.1109/HUMANOIDS65713.2025.11203198}{Kwon et al., 2025}\\[1pt]
  ALT \taxcite{https://proceedings.iclr.cc/paper_files/paper/2026/hash/122ea6470232ee5e79a2649243348005-Abstract-Conference.html}{He et al., 2026}
};
\node[papers] (learnpapers) at (10.75,-3.2) {
  HiRT \taxcite{https://proceedings.mlr.press/v270/zhang25b.html}{Zhang et al., 2025}\\[1pt]
  Realtime-VLA FLASH \taxcite{https://arxiv.org/abs/2605.13778v1}{Niu et al., 2026}\\[1pt]
  ACT (policy reference) \taxcite{https://roboticsproceedings.org/rss19/p016.html}{Zhao et al., 2023}
};
\draw[arr] (incall.east) -- (insidepapers.west);
\foreach \family/\paper in {horizon/horizonpapers,reuse/reusepapers,retrieve/retrievepapers,learn/learnpapers}
  \draw[arr] (\family.east) -- (\paper.west);
\end{tikzpicture}
\caption{\textbf{Two objectives and five families of VLA inference acceleration methods.}
In-call computation reduction forms one family; full-call reduction branches
into the other four. ACT is a standard lightweight-policy
reference, adapted in our comparison rather than originally proposed as a VLA
accelerator. Author--year labels link to verified primary records.}
\label{fig:teaser}
\end{figure}

Figure~\ref{fig:teaser} places the five families in Section~\ref{sec:related}
under two objectives: reducing computation within each VLA call and reducing
the number of full VLA calls. The latter has four branches: adaptive execution
chunk length, action reuse and extrapolation, action retrieval, and cheaper
learned policies.

\subsection{Reduce computation within a VLA call}
This family reduces the cost of one invocation through low-bit
execution \citep{zhang2026quantvla}, token pruning or reuse
\citep{liu2026vlapruner,xu2025vlacache}, adaptive depth
\citep{yue2024deervla}, latent prediction \citep{liu2026latentbridge}, and
post-backbone action caching \citep{oi2026actioncache}. These methods lower
selected terms inside a call, but the remaining terms impose an
Amdahl-style ceiling. They can complement VLA-ULAP by reducing computation
in the remaining VLA calls.

\subsection{Adaptive execution chunk length}
This family builds on action chunking, in which a policy predicts a sequence of
future actions in one call \citep{zhao2023act}. AutoHorizon reads the model's
action self-attention to estimate its predictive limit
\citep{wang2026autohorizon}; AAC uses action entropy to choose an execution
prefix \citep{liang2026aac}; DVAC chooses the low-denoising-variance prefix of
one flow-policy output \citep{feng2026dvac}; and A3 accepts the longest prefix
that remains conditionally consistent \citep{chen2026a3}. VLA-Corrector
truncates a chunk when latent visual dynamics indicate persistent deviation
\citep{pan2026vlacorrector}, while TempoWAM either continues the remaining
world-action-model chunk or replans \citep{ye2026tempowam}. These methods use
different reliability signals but select from the chunk already generated;
after that horizon, another policy call or a longer-horizon policy is required.

\subsection{Action reuse and extrapolation}
This family derives skipped actions from recent motion. FlashVLA, evaluated
on OpenVLA \citep{kim2024openvla}, repeats the previous action when recent
action directions and selected visual-token sets remain stable
\citep{tan2025flashvla}. SP-VLA fits a ridge regressor to a recent action buffer,
extrapolates the next action, and reuses the preceding gripper state
\citep{li2025spvla}. These paths are inexpensive, but their substitute action is
inherited from earlier motion rather than selected from the state that now
requires an action. Reuse and extrapolation can extend beyond the generated
chunk, but sustained reliance on earlier motion can sharply degrade success.
Our controlled SP-VLA comparison demonstrates this trade-off under aggressive
call reduction (Section~\ref{sec:ulap-spvla}). ULAP instead predicts a fresh
chunk conditioned on current observations, state, and executed action history.

\subsection{Retrieve executable actions}
Action retrieval instead returns an executable sequence from memory without
running the VLA on that decision. RT-Cache encodes the current RGB frame with
frozen DINOv2 \citep{oquab2024dinov2} and SigLIP \citep{zhai2023siglip},
concatenates their features, and retrieves a multi-step trajectory snippet
\citep{kwon2025rtcache}. ALT trains a contrastive fusion encoder over the
current first-person end-effector view, third-person view, and end-effector
pose; cosine lookup returns the demonstration chunk linked to the nearest
trajectory frame, while a similarity threshold flags unsupported observations
\citep{he2026alt}.

Both methods face limited cache support. RT-Cache reports 0\% success when no
in-domain frame exists for the tested object--camera--pose combination, and
image-key near-misses under occlusion, pose shifts, clutter, or scene dynamics
\citep{kwon2025rtcache}. ALT can detect an unsupported query and trigger a safe
fallback, but does not supply the correct action chunk for that OOD state
\citep{he2026alt}. RT-Cache also runs two frozen foundation vision encoders for
every query and searches a 2,176-dimensional index. In contrast, ULAP predicts
a new action chunk from observations and executed action history rather than
selecting a stored chunk. VLA-ULAP interleaves these local predictions with
calls to the unchanged base policy, using the VLA's broader generalization
capabilities to help maintain task success under out-of-distribution conditions.

\subsection{Learn a cheaper action policy}
HiRT combines a slowly updated VLM representation with a fast policy that
conditions on observations and the cached latent \citep{zhang2024hirt}.
When the VLM is remote, refreshing that latent requires feature transfer.
Realtime-VLA FLASH uses a draft model coupled to the main image encoder and
Action Expert verification \citep{niu2026realtimeflash}, rather than a fully
independent local path. In contrast, ULAP predicts complete chunks without
base-VLA hidden states or online verification and is trained independently
of the frozen VLA.

ACT predicts action chunks from images and robot state through imitation
learning \citep{zhao2023act}. It is not originally a VLA-call scheduler;
we adapt it as a standard lightweight action-chunk predictor under the same
hybrid schedule as ULAP, testing the local model rather than a different
call-allocation rule (Section~\ref{sec:ulap-spvla}).

\clearpage
\begingroup
\raggedbottom
\section{ULAP training and evaluation settings}
\label{app:ulap-fixed}

\paragraph{Predictor architecture.}
Table~\ref{tab:ulap-architecture} specifies the shared predictor blocks.
Let $H$, $D_a$, and $D_p$ denote the executed chunk length, action dimension,
and proprioception dimension. Their values are $(16,7,8)$ for GR00T,
$(7,7,8)$ for VLA-JEPA, $(16,7,9)$ for Cosmos-Policy, $(16,6,6)$ for SO-101,
and $(5,7,8)$ for the LIBERO-Safety predictor. The benchmark subsections
below specify the camera and task-conditioning inputs.

\begin{table}[htbp]
\centering
\caption{\textbf{ULAP predictor specification.} Linear layers include bias;
dropout is disabled at inference. The frozen visual encoder is Theia-Tiny.}
\label{tab:ulap-architecture}
\begin{tabular}{@{}p{0.22\linewidth}p{0.73\linewidth}@{}}
\toprule
Block & Structure \\
\midrule
Input projections & Visual tokens: LayerNorm then $192\to192$ linear.
Actions and state: $D_a\to192$ and $D_p\to192$ linear.
Learned position/type embeddings and a 192-D task vector. \\
Transformer & 3 pre-LayerNorm layers, width 192, 8 attention heads,
FFN width 768, GELU, dropout 0.15; final LayerNorm.
Aggregation-token projection $192\to128$, then $\ell_2$ normalization. \\
State MLP & $(HD_a+D_p)\to512\to512\to128$;
GELU after the first two layers, dropout 0.15 after the first GELU,
then $\ell_2$ normalization. \\
Fusion & Concatenate both 128-D outputs and divide by $\sqrt{2}$ (256-D). \\
Action head & Concatenate each learned 192-D step token with the fusion vector;
LayerNorm(448), $448\to192$, GELU, dropout 0.15, $192\to D_a$.
All $H$ steps share this head. \\
\bottomrule
\end{tabular}
\end{table}

Cosmos-Policy additionally projects its precomputed 1024-D language vector
through LayerNorm and a $1024\to192$ linear layer and adds it to the
task-conditioned aggregation token.

\paragraph{Optional ensemble.}
We use two predictors by default (four for CALVIN, Appendix~\ref{app:calvin-ulap}), with the
same architecture and optimization settings. Diversity comes from
repartitioning training/validation episodes within each task while preserving
their counts; all samples from an episode remain in the same partition.
The test set is unchanged. Each model computes normalization statistics
from its own training partition and selects its checkpoint using validation
macro-task standardized RMSE.

The predictors receive the same inputs and share frozen Theia features
computed once per local decision. Each action chunk is denormalized using
its model-specific statistics and clipped to the action bounds before
equal-weight FP32 averaging across all steps and action dimensions,
including the gripper. The averaged actions are executed and supply the
subsequent action history. The base VLA, initial VLA call, and fixed schedule
remain unchanged. Additional computation consists of lightweight
predictor passes and action averaging, without repeating visual encoding.

\subsection{LIBERO $\times$ GR00T N1.7}
\label{app:groot-ulap-fixed}

\paragraph{Evaluation.}
We use NVIDIA GR00T-N1.7-LIBERO on four suites of ten tasks each.
Every condition evaluates official initial-state IDs 0--49 per task
(2,000 episodes), with environment seed 7 and policy seeds 7--56.
Both paths execute 16-step chunks; the episode limit is 720 control steps.
The fixed scheduler evenly spreads local decisions, accounting for the mandatory
first VLA chunk. Requested local fractions are 0.2--1.0 in increments of 0.1,
plus the VLA-only reference. All nine settings use
the same suite-specific checkpoints and initial-state protocol.

\paragraph{Training.}
Suite-specific predictors are trained on successful GR00T rollouts collected
with a 100-rollout budget per task, using episode-level splits and
training-only normalization. Training uses
Smooth-L1 ($\beta=1$), AdamW (learning rate $6\times10^{-4}$, weight decay 0.05),
batch 96, 120 epochs, eight warmup epochs followed by cosine decay,
bf16 training autocast with FP32 parameters and validation, and seed 0;
validation selects the checkpoint without refitting.
Inputs are current primary/wrist images, proprioception, executed action history,
and task conditioning.
Predictor-training initial states are disjoint from evaluation.

\paragraph{Input ablations (Appendix~\ref{app:ulap-input-ablations}).}
Both variants retain current images and task conditioning and use the same data,
optimizer, training seed, checkpoint selection, and evaluation protocol as the
full model. Removing the state MLP retains history and proprioception in the
Transformer; removing action history removes it from both branches while
retaining proprioception. Each of the nine fixed-schedule settings evaluates
2,000 episodes against the same VLA-only reference (1,863 successes; 24,466 calls).

\paragraph{Metrics (simulation evaluations).}
Success rate is the fraction of evaluation episodes that succeed. VLA-call
reduction compares total VLA calls across all evaluation episodes, including
both successful and failed trials, against the VLA-only baseline:
$1-\sum_i C_i^{\mathrm{hybrid}}/\sum_i C_i^{\mathrm{exact}}$. This is not the local-decision
fraction. It captures the realized reduction in total VLA calls during execution,
including changes in episode length.

\begin{figure}[H]
\centering
\includegraphics{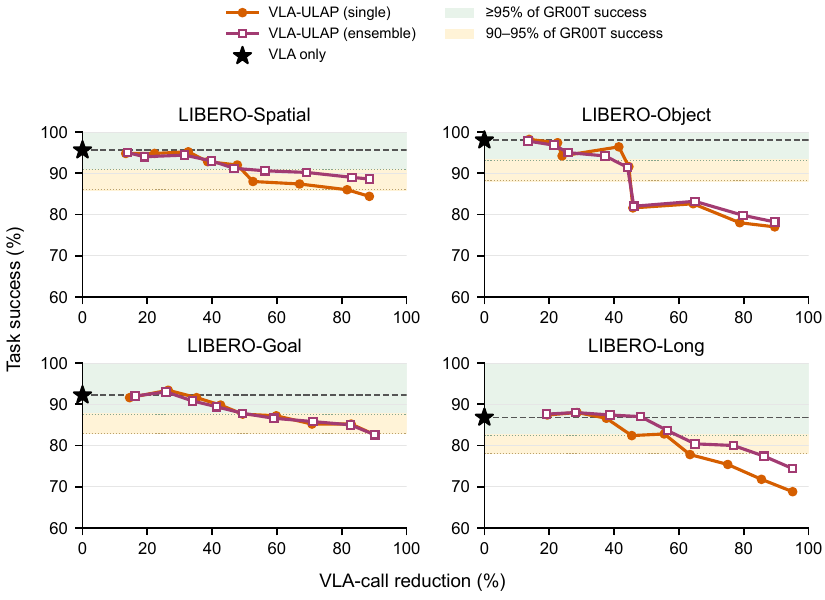}
\caption{\textbf{GR00T N1.7 with ULAP by LIBERO suite.} Each point covers
500 episodes. Stars/dashed lines mark the baseline success rate. Green and amber bands indicate
$\geq$95\% and 90--95\% retention. Orange circles show the single predictor;
raspberry squares show the two-model ensemble. Both use fixed scheduling.}
\label{fig:ulap-fixed-suites}
\end{figure}

\paragraph{Ensemble results.}
The two-model ensemble uses the same nine local fractions and paired initial
states as the single predictor. At local fraction 1.0, success increases from
78.20\% to 80.95\%, with both methods reducing VLA calls by 91.83\%.
The corresponding gains on LIBERO-Long, Object, Goal, and Spatial are
5.60, 1.20, 0.00, and 4.20~pp, respectively.

\clearpage
\subsection{LIBERO $\times$ VLA-JEPA}
\label{app:jepa-ulap-fixed}

\paragraph{Evaluation.}
We use the released VLA-JEPA LIBERO checkpoint with its state-conditioned
action head, ten DDIM steps, and Franka action normalization.
Each condition evaluates the same 40 tasks and official initial-state IDs
0--49 (2,000 episodes), with environment seed 7 and ten settling steps.
Both paths execute seven-step chunks. Episode limits are 250, 280, 300, and
520 control steps for Spatial, Object, Goal, and Long, respectively.
VLA-call seeds depend on the base seed, suite, trial, and control-step
index, independently of the requested local fraction. The fixed schedule
and requested fractions match Appendix~\ref{app:groot-ulap-fixed}.
ULAP predicts actions directly in bf16, without running the VLA action head.

\paragraph{Training.}
Four suite-specific predictors are trained on successful VLA-JEPA rollouts: 3,912
trajectories from 100 attempts per task. The recorded simulator-state audit
confirms no overlap with evaluation initial states ($L_\infty$ tolerance
$10^{-4}$). Examples pair current primary/wrist camera views,
current proprioception, and the previous executed chunk with the next
action chunk. Task-local episode splits are 70/15/15\% for training,
validation, and test, with training-only normalization.
Optimization uses the same loss, AdamW settings, batch size, epoch count,
warmup/cosine schedule, and bf16 precision as GR00T, but training seed 7.
Validation selects one checkpoint per suite, without refitting.
The predictor has 1,871,815 trainable parameters, excluding frozen Theia.
Previous images are omitted: 98 visual tokens and nine action/state/aggregation
tokens give 107 input tokens. The four predictors are trained from initialization
with the same recipe on their respective suites' rollouts.

\paragraph{Ensemble.}
The two-model ensemble follows the shared setup in Appendix~\ref{app:ulap-fixed}.
For each suite, the second predictor uses an episode-level training/validation
repartition (split seed 20260923), training seed 0, 120 epochs, and batch size 96.
The test split is excluded, normalization uses training data only, and validation
selects the checkpoint without refitting. The averaged action chunk is passed
through the evaluator's usual gripper binarization. We evaluate the same
initial states at nine local fractions from 0.2 to 1.0.

\begin{figure}[H]
\centering
\includegraphics{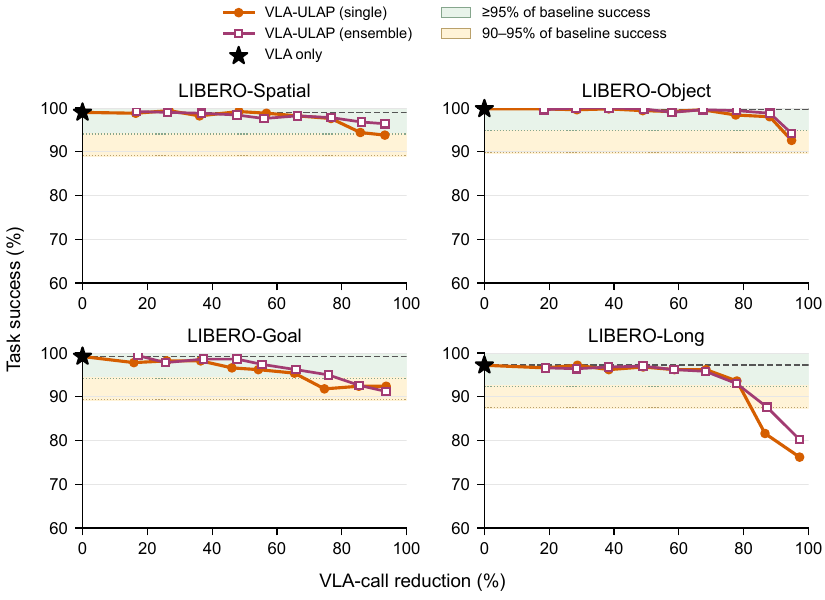}
\caption{\textbf{VLA-JEPA with ULAP by LIBERO suite.} Each point covers
500 episodes. Each panel uses its own VLA-only call total and baseline success rate
(star/dashed line). Green and amber bands indicate $\geq$95\% and 90--95\%
retention. Filled circles denote single predictors; open squares denote ensembles.}
\label{fig:jepa-ulap-fixed-suites}
\end{figure}

\paragraph{ACT, SP-VLA, and RT-Cache comparison.}
\label{app:jepa-spvla}
All methods use the same 40 tasks, 50 initial states per task,
environment seed, and episode limits as above; the hybrid methods share
the same VLA checkpoint.
ACT~\citep{zhao2023act} uses four suite-specific validation-selected policies
with current front/wrist images, proprioception, and task identity.
ACT is trained on the same successful VLA-JEPA rollout corpus and episode
splits as ULAP, with the same 120-epoch budget.
The LeRobot ACT implementation uses an ImageNet-initialized ResNet18,
a width-512 Transformer, and CVAE training. Inference uses FP32 and no temporal
ensembling, with the CVAE latent variable $z$ fixed to its prior mean of zero
rather than sampled. ACT replaces the local branch under the same
fixed scheduler at fractions 0.2/0.5/0.8/0.9/1.0, generating and executing
seven actions per query. Even at 1.0, the initial query uses the VLA.
This is our ACT-based bypass baseline, not the original ACT evaluation protocol.

SP-VLA's scheduling component~\citep{li2025spvla} uses seven-step exact chunks; local ridge
extrapolation generates one action from six previous actions
($\lambda=10^{-4}$). We disable token pruning to focus on reducing complete
VLA calls rather than computation within each call.
The three plotted settings comprise the official freshness/motion gates
and two experimental high-reduction variants: motion thresholds scaled
by $\times8$ and $\times32$, with consecutive extrapolation capped at 4 and 8 steps, respectively,
replacing the official requirement for four exact actions among the last six.
These variants were calibrated on 160 successful training/validation
trajectories disjoint from evaluation, then frozen.

RT-Cache~\citep{kwon2025rtcache} retrieves from task-specific memories built
from the same corpus's 2,740 training-split successful teacher rollouts.
It encodes the current primary-camera image with frozen DINOv2 and SigLIP,
normalizes the concatenated 2,176-D feature, and averages the top-five
chunks under exact cosine similarity. We sweep chunk lengths 4/6/8/12/16,
constructed from executed teacher actions. Retrieval starts at the first
decision, without VLA calls or fallback. This LIBERO implementation uses
in-memory search rather than RT-Cache's database-serving stack.

\paragraph{Successful-episode inference costs.}
For each successful episode $i$, we estimate
$T_i=N_{V,i}t_V+N_{L,i}t_L$ and $E_i=N_{V,i}e_V+N_{L,i}e_L$.
Here, $T_i$ and $E_i$ are the episode's estimated total inference time and
GPU-board energy. $N_{V,i}$ and $N_{L,i}$ count VLA and local-path invocations
in episode $i$, respectively. $t_V,t_L$ are the measured mean times per
invocation, and $e_V,e_L$ are the corresponding mean GPU-board energies.
The local path is ULAP, ACT, SP-VLA extrapolation, or RT-Cache retrieval.
For standalone RT-Cache, $N_{V,i}=0$ and local costs are task- and chunk-specific.
We then average $T_i$ and $E_i$ over that method's successful episodes. Counts include the
initial query and any partially executed final chunk.
Batch-one RTX A6000 profiles give 212.195\,ms/27.3115\,J per VLA query,
24.503\,ms/2.7668\,J per ACT query, and 17.043\,ms/1.6741\,J per ULAP
query including Theia. SP-VLA costs 0.176--0.177\,ms and
0.00402--0.00417\,J per extrapolated action, using the stable final two
profiling blocks. ULAP uses the two current camera views; its profile pools
7,011 calls across six blocks in two sessions, each following 300 warm-up calls.

Energy includes gross GPU-board consumption during inference, including
resident idle power for CPU-based SP-VLA, but excludes CPU/host energy,
simulation, communication, server queuing, and route-switch transients.

\clearpage
\subsection{RoboCasa $\times$ Cosmos-Policy}
\label{app:cosmos-ulap-fixed}

\paragraph{Evaluation.}
We use NVIDIA Cosmos-Policy-RoboCasa-Predict2-2B in bf16 with TF32 disabled
and five denoising steps. The base policy generates 32 actions and executes
the first 16; ULAP predicts and executes 16 actions. Each condition evaluates
24 tasks across five (layout, style) pairs, $(1,1)$, $(2,2)$, $(4,4)$, $(6,9)$, and
$(7,10)$, with ten episodes per pair (1,200 episodes).
Object split B is used throughout evaluation. Environment seeds vary by
task and episode, while the VLA seed is 195. The same identities, initial-state
hashes, and object hashes are matched across conditions. Ten settling steps
precede control, which terminates on success or at the official task-specific
step limit. We use the fixed scheduler with requested local
fractions $0.2,0.3,\ldots,1.0$, always starting with one VLA call.
All nine fractions use the same checkpoint and 1,200 initial conditions.

\paragraph{Training.}
A single predictor covers all 24 tasks, using current primary,
secondary, and wrist views, nine-dimensional proprioception, executed action
history, and precomputed frozen language embeddings. Successful teacher
rollouts from object split A provide 106,061 training examples at stride four;
validation and test splits are separated by collection episode, with per-task
training-only normalization. We use standardized Smooth-L1 loss ($\beta=1$),
AdamW (learning rate $6\times10^{-4}$, weight decay 0.05), global batch size 192,
task-balanced sampling, 120 epochs, eight warmup epochs followed by cosine
decay, bf16, and seed 0. The predictor has 2,118,791 trainable parameters, excluding frozen
encoders.
The Transformer input consists of 165 tokens.

\paragraph{Task-level results.}
Figure~\ref{fig:cosmos-ulap-tasks} shows single-predictor and ensemble results
across all 24 tasks; the ensemble covers requested local fractions 0.2--1.0.
For the single predictor, at requested local fraction 0.5, success is
769/1,200 (64.08\%), or 97.47\% of baseline, with 48.79\% fewer VLA calls.
At local fraction 0.6, the higher reduction of 56.69\% retains 91.76\% of baseline
success (724/1,200).
The ensemble also preserves success across individual tasks. With an operating
point selected separately for each task, it retains at least 90\% of VLA-only
success while reducing VLA calls by at least 26.63\% on every task,
including reductions of at least 40\% on 19 of 24 tasks.
The small success gain at local fraction 0.2 is a point
estimate, not evidence of general superiority over the base policy.
\begin{figure}[p]
\centering
\includegraphics{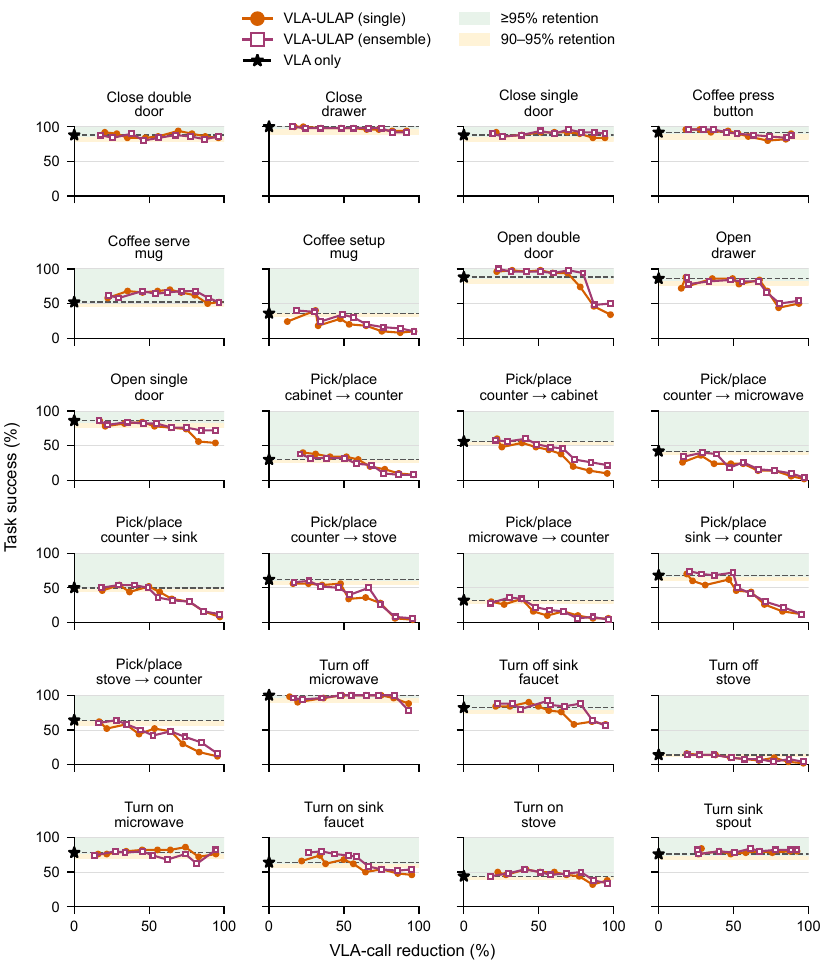}
\caption{\textbf{RoboCasa success across all 24 tasks.} Each point uses 50
matched episodes. The black star and dashed line mark VLA-only
success; bands indicate $\geq95\%$ and 90--95\% retention. Orange circles
show the single predictor and raspberry squares the two-model ensemble,
both with fixed scheduling. Call reduction is measured against the
corresponding task baseline.}
\label{fig:cosmos-ulap-tasks}
\end{figure}

\clearpage
\subsection{CALVIN $\times$ X-VLA}
\label{app:calvin-ulap}

\paragraph{Motivation.}
We use CALVIN~\citep{mees2022calvin} to evaluate long-horizon, multi-task
execution and generalization across environments. Five language instructions
are executed consecutively without resetting the environment, so each task
inherits the state and any prediction errors left by earlier tasks.
Predictors trained on environments A, B, and C are evaluated in D, which
changes scene textures and the positions of fixtures such as drawers, sliding
doors, and switches. This tests transfer of known manipulation skills to a
different visual appearance and spatial layout. We pair ULAP with
X-VLA~\citep{zheng2026xvla} and use a four-model ensemble to improve robustness
against error accumulation over these longer chains.

\begin{figure}[H]
\centering
\includegraphics{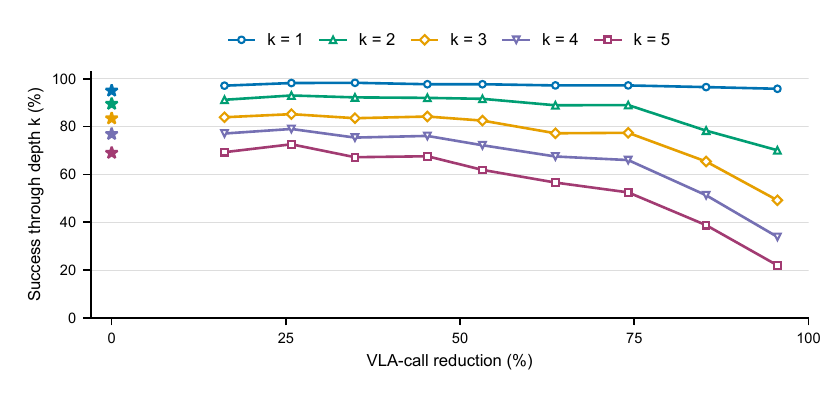}
\caption{\textbf{CALVIN results by instruction depth.}
Success rate for completing the first $k$ instructions consecutively,
with one curve per depth $k=1,\ldots,5$. VLA-call reduction uses all
1,000 chains. Stars mark VLA-only baselines in matching depth colors.}
\label{fig:calvin-ulap-depth}
\end{figure}

\paragraph{Nearly half the calls removed with strong chain completion.}
At requested local fraction $\rho=0.5$, VLA-ULAP reduces VLA calls by
\textbf{45.3\%} while retaining \textbf{98.0\% of the baseline five-task success rate}
(Figure~\ref{fig:calvin-ulap-tradeoff}). Mean completed tasks do not decrease,
reaching 4.176 versus 4.138 for VLA only.

Figure~\ref{fig:calvin-ulap-depth} shows that at $\rho=0.5$, even the
longest chains incur only a small success-rate decrease, at most 1.4~pp
across all depths. With more aggressive skipping, however, performance
declines more strongly as the number of chained tasks grows, consistent
with prediction errors accumulating during extended local execution.
These results support substantial call reduction in long-horizon execution
while identifying the growing importance of VLA calls as chains lengthen.
Each lightweight predictor also shares its parameters across all 34 task
categories, using language to select the skill rather than requiring a
separate network per task. The ensemble combines four such shared predictors.

\paragraph{Training and evaluation settings.}
We evaluate the same 1,000 official five-instruction chains in environment D
with the frozen X-VLA checkpoint fine-tuned on ABC. Each instruction has a
360-control-step budget at 30~Hz; a timeout ends the chain. Both paths execute
20-step chunks. The fixed scheduler starts with a VLA call and retains its
state across instruction changes, while discarding any unexecuted actions
for the previous instruction.

For chain $i$, let $L_i\in\{0,\ldots,5\}$ be the number of instructions
completed consecutively before failure or full completion. At instruction
depth $k$, success means completing all of the first $k$ instructions.
With $N=1{,}000$ evaluated chains, its rate and the mean completed-task count are
\[
S_k=\frac{1}{N}\sum_{i=1}^{N}I(L_i\geq k),
\qquad
\overline{L}=\frac{1}{N}\sum_{i=1}^{N}L_i=\sum_{k=1}^{5}S_k,
\]
where the indicator $I(\cdot)$ equals one when its condition holds and zero otherwise.
Figure~\ref{fig:calvin-ulap-depth} plots $100S_k$ for each $k$.
Every depth uses all $N$ chains, including those that fail earlier.
Call reduction likewise uses total VLA calls over all chains, relative
to the matched VLA-only baseline.

Each ensemble member is trained on successful subtask segments from 960 ABC
rollout sequences, with 120 additional sequences for validation. The four
members use different trajectory-level splits of the same 1,080-sequence
pool. No D trajectories enter predictor training. Current static and wrist
views, proprioception, and 20 executed actions are supplemented by a frozen
384-dimensional language embedding, cached per instruction. The shared
Theia-Tiny encoder supplies $7\times7$ tokens per view. Each member has
2.01M trainable parameters across a three-layer, width-192 Transformer,
a state MLP, and a flat one-pass chunk head. Training uses masked Smooth-L1 loss and
AdamW for 120 epochs, batch size 96, learning rate $6\times10^{-4}$,
and eight warmup epochs followed by cosine decay. Decoded position and
gripper predictions are averaged; averaged rotations are projected onto
$\mathrm{SO}(3)$.

\clearpage
\subsection{Real robot $\times$ GR00T N1.7}
\label{app:real-robot-ulap}

\paragraph{Protocol.}
Each task uses a task-specific GR00T N1.7 checkpoint fine-tuned on 100 human
demonstrations from ten object placements (P1--P10), selected at training
step 2,000. The paired evaluation uses five trials at each P1--P10 placement
and five at each held-out H1--H5 placement, in the same frozen trial order
for both policies. OOD here denotes unseen object placements, not unseen
object categories or tasks. Control is synchronous at a target 30\,Hz, with
a common fixed home pose, unchanged safety clamps, and a 60-second timeout.
Success requires the object to remain in its target container, released and
clear of the gripper, for two seconds. Both paths execute 16 actions per
decision. The first decision uses GR00T, followed by the fixed evenly spread
schedule with requested local fraction 0.75 for both tasks.
Only final valid trials after infrastructure-failure recollection are counted.

\paragraph{Predictor and training.}
Frozen Theia features from current and previous external/wrist images are
fused with the measured six-dimensional state and the previous 16 executed
actions. Normalized Transformer and state-MLP branches have fixed equal
weight; a parallel temporal head predicts a $16\times6$ action chunk.
The five body axes are relative to the current measured state, while the
gripper target is absolute. Each task's 100 human demonstrations are split
80/10/10 by episode for training/validation/test, with stride-four examples
and training-only normalization. Training uses standardized Smooth-L1
loss ($\beta=1$), AdamW (learning rate $6\times10^{-4}$, weight decay 0.05),
batch 96 and 120 epochs, with eight warmup epochs and cosine decay.
Two independently trained predictors (seeds 7 and 0) share frozen Theia features.
Each member decodes and clips its absolute actions before their equal-weight
mean is executed, including the gripper channel.

\paragraph{Cloud VLA-call reduction.}
For efficiency and safety, clearly failed physical runs were stopped early.
We therefore compare mean cloud VLA calls per successful episode under
VLA-ULAP and GR00T-only, respectively:
$R=1-\overline{C}_{\mathrm{ULAP},\,\mathrm{success}}/
\overline{C}_{\mathrm{GR00T},\,\mathrm{success}}$.
These calls invoke the base VLA on the remote GPU.
Ping-pong uses 1,088 cloud VLA calls across 63 baseline successes and 359 across
70 VLA-ULAP successes; Glue-stick uses 1,368 across 72 and 390 across 73, respectively.
Table~\ref{tab:physical-single-ensemble} compares single-predictor and ensemble
results. Despite using a higher requested local fraction (0.75 versus 0.60
and 0.67), the ensemble completes more trials successfully on both tasks.

\begin{table}[!htbp]
\centering
\caption{\textbf{Single and ensemble physical evaluations.} Cloud VLA calls are means per successful episode.}
\label{tab:physical-single-ensemble}
\begin{tabular}{llrrrr}
\toprule
Task & Policy & Local fraction & ID & OOD & Cloud VLA calls \\
\midrule
Ping-pong & GR00T & 0 & 41/50 & 22/25 & 17.27 \\
 & Single & 0.60 & 40/50 & 20/25 & 8.23 \\
 & Ensemble & 0.75 & 48/50 & 22/25 & 5.13 \\
Glue-stick & GR00T & 0 & 47/50 & 25/25 & 19.00 \\
 & Single & 0.67 & 48/50 & 24/25 & 7.07 \\
 & Ensemble & 0.75 & 48/50 & 25/25 & 5.34 \\
\bottomrule
\end{tabular}
\end{table}

\paragraph{A6000--Jetson cost model.}
\label{app:ulap-jetson-cost}
We profile the SO-101 Glue-stick GR00T checkpoint on RTX A6000 at batch one,
with four denoising steps and its bf16-compute/FP32-parameter configuration.
The path includes image packing, preprocessing, model inference, action
decoding and CPU output of 16 actions. GR00T computes its native 40-position
horizon and returns the first 16 actions. Both devices use the same saved
Glue-stick validation query. After 30 warmup calls and thermal stabilization,
300 synchronized calls average 289.27\,ms for GR00T and 20.72\,ms for ensemble ULAP.
Jetson Orin Nano uses 15\,W DVFS, FP32/TF32 CUDA Graphs, shared Theia and
separate streams for the two predictor members.

Energy is idle-subtracted on both devices. We integrate power over a separate
continuous-inference window and subtract resident-model idle power measured
before and after it. For active duration $T$ and $N$ completed calls,
$e=(E_{\mathrm{active}}-P_{\mathrm{idle}}T)/N$.
This gives 40.462\,J per GR00T call from NVML GPU-board power and 0.122\,J
per ensemble ULAP call from Jetson VDD\_IN module power (CPU, GPU and DRAM).
Table~\ref{tab:jetson-single-ensemble} compares single and ensemble inference
under the same DVFS configuration and saved input. Sharing frozen vision
keeps the ensemble's full-path latency overhead to 5.0\% and energy overhead to 7.1\%.

\begin{table}[!htbp]
\centering
\caption{\textbf{Measured Jetson Orin Nano costs.} Latency is the mean of 300 calls;
energy is idle-subtracted per completed inference in a separate continuous window.}
\label{tab:jetson-single-ensemble}
\begin{tabular}{lrrr}
\toprule
Local predictor & Mean latency (ms) & Energy (J) & Active / idle power (W) \\
\midrule
Single & 19.73 & 0.1140 & 11.01 / 5.69 \\
Ensemble (default) & 20.72 & 0.1221 & 11.12 / 5.68 \\
\bottomrule
\end{tabular}
\end{table}

For each policy, we average cloud VLA and local ULAP invocation counts over
its successful episodes and use
$\widehat C=\overline N_{\mathrm{VLA}}c_{\mathrm{A6000}}+
\overline N_{\mathrm{ULAP}}c_{\mathrm{Jetson}}$.
Here, $\overline N_{\mathrm{VLA}}$ and $\overline N_{\mathrm{ULAP}}$ are the mean
numbers of cloud VLA calls and local ULAP predictions per successful episode, respectively.
$c_{\mathrm{A6000}}$ and $c_{\mathrm{Jetson}}$ are the measured time or energy
per call on the respective devices, and $\widehat C$ is the estimated mean
total inference time or energy per successful episode.
Camera acquisition, communication, queueing and actuation are outside the estimate. Energy also
excludes the A6000 host and subtracts device idle energy within active windows.

\paragraph{Illustrative deployment profiles.}
\label{app:deployment-profiles}
To illustrate the broad differences in compute capability, power range, and
price between the two devices, Table~\ref{tab:deployment-devices} summarizes
their published specifications. This complements the measured application
costs in Figure~\ref{fig:ulap-jetson-cost}.
\begin{table}[!htbp]
\centering
\caption{\textbf{Remote GPU and local edge-device specifications.}
NVIDIA's A6000 datasheet and Jetson announcement
\citep{nvidia2022a6000,sheshadri2024orinsuper}.
Peak compute uses different precision/sparsity and is not a speedup comparison.
The A6000 price is an indicative GPU-only
\href{https://www.newegg.com/pny-technologies-inc-vcnrtxa6000-pb-rtx-a6000-48gb-graphics-card/p/N82E16814133822}{retail quote}
(September 2026); the Jetson price is the announced developer-kit offer.
Our ULAP measurements use 15 W DVFS, not the
highest Super power mode.}
\label{tab:deployment-devices}
\begin{tabular}{@{}p{1.40in}p{1.64in}p{2.08in}@{}}
\toprule
& RTX A6000 & Jetson Orin Nano Super (8GB) \\
\midrule
Role in VLA-ULAP & Remote VLA GPU & Local ULAP device \\
Memory & 48 GB GDDR6 & 8 GB LPDDR5 \\
Memory bandwidth & 768 GB/s & 102 GB/s \\
Published peak compute & 38.7 TFLOPS (FP32) & 67 TOPS (sparse INT8) \\
Specified power & 300 W maximum board & 7 / 15 / 25 W module modes \\
Indicative price & $\sim$US\$6,400 (GPU only) & US\$249 (kit, December 2024) \\
\bottomrule
\end{tabular}

\end{table}

Figure~\ref{fig:ulap-deployment} uses the measured physical Glue-stick profiles
from the A6000--Jetson cost model above. Both paths output $16\times6$ action
chunks. GR00T N1.7 on RTX A6000 takes 289.3\,ms and 40.46\,J per inference,
while ensemble ULAP on Jetson Orin Nano takes 20.7\,ms and 0.122\,J.
Energy covers the GPU board for A6000 and the VDD\_IN module for Jetson,
including its CPU, GPU, and DRAM; both values subtract resident-model idle power.
Energy per inference is obtained from continuous-inference windows, rather
than multiplying separately profiled latency by power.

The communication scale in Figure~\ref{fig:ulap-deployment} follows
DaDu-Corki's RoboFlamingo setup, with a Franka Panda communicating with a
V100 server over Wi-Fi \citep{huang2025daducorki}. Figure~2 of \citet{huang2025daducorki} shows
image-transfer latency of tens of milliseconds and energy of hundreds of
millijoules per frame. These are coarse visual readings of its plot.
The study measures actual frame transfers and integrates measured
power over time. These communication-stage values are not a separately
verified observation-upload/action-return round trip, and do not enter our
inference-device cost estimates.
Figure~\ref{fig:ulap-deployment} reports the mean success rate over the two
dynamic tasks, the Glue-stick successful-episode sums of VLA and ULAP inference
costs, and the measured per-inference branch costs described above. Its episode
costs use both paths, not just avoided VLA calls. Battery-life benefits concern
low-power ULAP deployment instead of carrying a full-VLA GPU onboard.
The drawing's robot counts illustrate the capacity opportunity, not measured
server limits. With fewer requests per robot, a shared service can devote
more compute time to other clients; the feasible robot count also depends on
request deadlines, queueing, batching, and the set of resident policies.
The affordable edge device and reduced per-robot inference demand support
battery-constrained robots and shared-GPU fleets.

\subsection{LIBERO-Safety $\times$ $\pi_{0.5}$}
\label{app:ulap-safety}

\paragraph{Latency-aware evaluation.}
We use the LIBERO-Safety-fine-tuned $\pi_{0.5}$ checkpoint and the L1
\texttt{obstacle\_avoidance} tasks 2 and 4, selected by an earlier latency
sweep over the suite's five tasks. Each evaluation condition uses the same 50 official
initial states four times, with seed base 195 and a 600-step limit.
The VLA generates ten actions; both paths execute five per decision.
The first decision always invokes the VLA, followed by a fixed 50\% local
schedule. At 20\,Hz, one control period is 50\,ms. Batch-one measurements on
RTX A6000 give mean request--response latencies of 18.0--18.2\,ms for ULAP
and 147.5\,ms for $\pi_{0.5}$ including client preprocessing.
These correspond to less than one and approximately three control periods,
respectively. The evaluation uses three injected delay steps for the VLA
and zero additional steps for local prediction.
Inference is issued before the current interval ends; execution
continues from the previous chunk during the delay, and the elapsed prefix
of the arriving chunk is skipped.

\paragraph{Predictor training.}
Separate predictors use successful, zero-delay teacher rollouts, with
368 Moka-pot and 418 Mug episodes split 70/15/15\% by episode for training,
validation, and testing. Collection initial states are separated from the
evaluation states by a configuration-space distance check.
This variant uses current camera views only, frozen Theia-Tiny features,
current proprioception, and the five previously executed actions.
Each predictor has 1,862,151 trainable parameters excluding Theia and produces
five actions in parallel. Training uses standardized Smooth-L1 loss,
AdamW (learning rate $6\times10^{-4}$, weight decay 0.05), batch size 96,
and up to 120 epochs with early stopping.

\clearpage
\section{Ablation studies}
\label{app:ulap-ablations}
\subsection{Input ablations on GR00T N1.7}
\label{app:ulap-input-ablations}

\label{sec:ulap-input-ablations}

We ablate the state MLP and executed-action history to assess ULAP's design
(Figure~\ref{fig:ulap-input-ablations}). Removing the state MLP lowers
success at all but the smallest local fraction. In most evaluated settings
with moderate VLA-call reduction ($\leq 50\%$) and high success rates,
action history yields higher success rates at comparable VLA-call reduction.
This suggests history is useful under limited distribution drift; its benefit
reverses under aggressive bypass, plausibly as local predictions shift histories
away from the training distribution. Retraining and evaluation settings are
specified in Appendix~\ref{app:groot-ulap-fixed}.

\begin{figure}[H]
\centering
\includegraphics{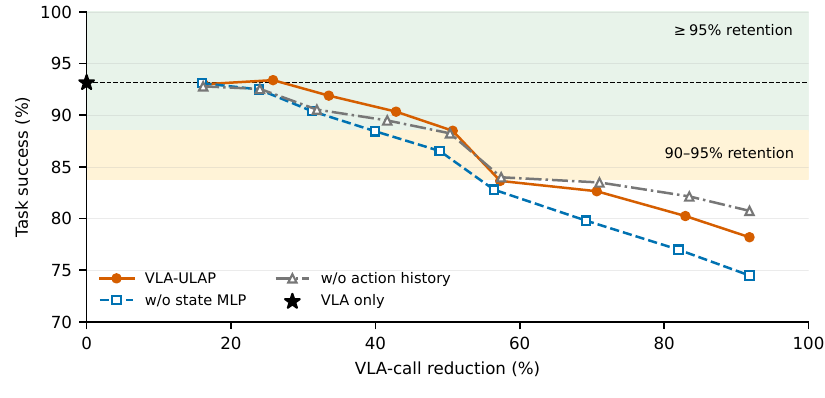}
\caption{\textbf{Input ablations on GR00T N1.7.} Colors and retention bands
follow Figure~\ref{fig:ulap-fixed}.}
\label{fig:ulap-input-ablations}
\end{figure}

The sweep connects requested local fractions
$\rho=0.2,0.3,\ldots,1.0$. VLA-only succeeds in 1,863/2,000 episodes
(93.15\%) with 24,466 calls.

\clearpage
\subsection{Vision-backbone scale on GR00T N1.7}
\label{app:ulap-theia-scale}

We replace frozen Theia-Tiny with frozen Theia-Small while retaining the
full-input ULAP design. The backbone grows from 5.52M to 21.67M parameters;
both use the same image resolution, patch size, and pooled token count.
Only the input normalization and projection widen to accommodate Small's
features; the remaining predictor architecture, training data, hyperparameters,
and validation-based checkpoint selection are unchanged. Both variants use
training seed 0 and the matched fixed-schedule evaluation in
Appendix~\ref{app:groot-ulap-fixed}.

\begin{figure}[H]
\centering
\includegraphics{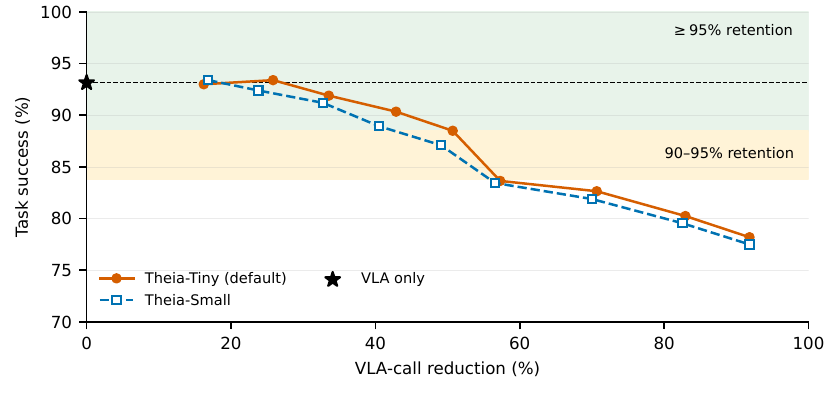}
\caption{\textbf{Scaling the frozen vision backbone from Theia-Tiny to Theia-Small.}
Each point covers 2,000 matched episodes across all 40 LIBERO tasks, with
16-step chunks and requested local fractions $\rho=0.2,0.3,\ldots,1.0$.
VLA-call reduction includes failures and uses the same VLA-only denominator
as Figure~\ref{fig:ulap-input-ablations}. Axes, baseline, and retention bands
follow that figure.}
\label{fig:ulap-theia-scale}
\end{figure}

\paragraph{Theia-Tiny is sufficient for this setting.}
Scaling to Small yields no statistically significant success difference at
any tested local fraction (two-sided exact McNemar tests, Holm correction
across the nine fractions; all adjusted $p>0.71$).
Tiny achieves higher success with at least as much VLA-call reduction at
eight of the nine settings (Figure~\ref{fig:ulap-theia-scale}). Thus, the
larger backbone offers no demonstrated benefit here, supporting Theia-Tiny
as the compact default for ULAP.

\clearpage
\subsection{Spatial visual-token resolution on GR00T N1.7}
\label{app:ulap-token-grid}

The backbone-scale ablation asks how much visual capacity ULAP needs;
we next test how much spatial detail it should preserve. We pool the same
frozen Theia-Tiny features to $1\times1$, $7\times7$, or the native
$14\times14$ grid per camera, giving 2, 98, or 392 visual tokens across two
views. Adding 16 action tokens, one proprioception token, and one
task-conditioned aggregation token gives 20, 116, or 410 tokens in total.
Only the visual positional-embedding size changes with the grid; all variants
retain the same training data, seed 0, 120-epoch training recipe, and
validation-based checkpoint selection. We evaluate 16-step chunks on the same
2,000 episodes per fixed-schedule setting as Appendix~\ref{app:groot-ulap-fixed}.

\begin{figure}[H]
\centering
\includegraphics{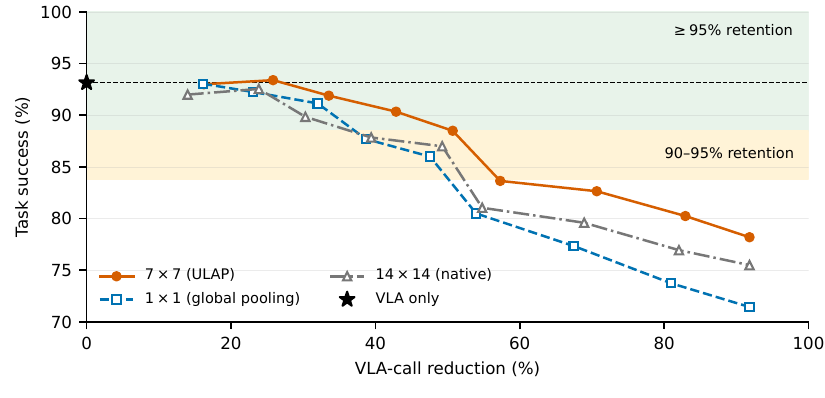}
\caption{\textbf{Spatial pooling balances detail and compactness.}
Each point covers 2,000 matched episodes across 40 LIBERO tasks.
Lines connect the nine fixed schedules, $\rho=0.2,0.3,\ldots,1.0$.
VLA-call reduction compares total calls over all episodes with the matching
VLA-only baseline. The star, dashed baseline and retention bands follow
Figure~\ref{fig:ulap-input-ablations}.}
\label{fig:ulap-token-grid}
\end{figure}

\paragraph{A compact spatial grid outperforms both extremes.}
The $7\times7$ grid achieves at least as much call reduction and higher
success than $1\times1$ at eight of nine settings, and than $14\times14$
at all nine settings (Figure~\ref{fig:ulap-token-grid}). At the highest
call reduction, 91.83\%, success is 78.20\% with $7\times7$, compared with
71.45\% for global pooling and 75.50\% for the native grid. Thus, collapsing
each view to one token loses useful spatial information, while retaining four
times as many visual tokens does not improve success under the same training
recipe. Together with the backbone-scale result, this supports a small frozen
encoder with a moderately pooled spatial representation, rather than either
discarding spatial structure or increasing visual capacity indiscriminately.

\paragraph{Why can the finer grid hurt?}
On Spatial, Object, and Goal, $14\times14$ yields lower training RMSE but
higher validation RMSE than $7\times7$, consistent with overfitting under
the shared training recipe. Pooling $2\times2$ neighborhoods removes
46.5--48.5\% of within-frame spatial variation in these suites. This suggests
that moderate pooling regularizes the predictor by suppressing fine-grained,
rollout-specific variation while retaining the spatial structure needed for
action prediction.

\clearpage
\section{Exploring adaptive cloud--local scheduling}
\label{app:ulap-cads}

\paragraph{Motivation.}
Fixed scheduling distributes VLA calls evenly, regardless of the current
scene or the local predictor's reliability. We investigate whether adaptive
cloud--local scheduling can achieve higher task success rates than fixed
scheduling by allocating VLA calls to decisions where they are more useful.
To keep the additional routing computation small, we use information already
available from observations and ULAP---visual features, predicted actions,
or predictor embeddings---rather than train another routing model.
We first compare three
signals offline, then examine the strongest candidate, risk-aware scheduling,
in closed-loop evaluation.

\paragraph{Candidate signals.}
We consider three signals for deciding when to prioritize a VLA call.
\emph{Image change} measures the change in Theia features over one executed
chunk. For each camera, we average-pool the $7\times7$ token grid to $2\times2$,
normalize each cell's feature, and average squared distances between matching
current and previous cells. \emph{Action change} is the RMS difference between
ULAP's predicted chunk and the last executed chunk, aligned by relative step
and standardized using training-set action statistics, including the gripper.
\emph{Embedding risk} measures how far ULAP's current representation lies from
same-task training representations, as defined below. None requires an extra
learned encoder. Image history is used only by the image-change scheduler,
not as an additional predictor input.

\paragraph{Offline comparison.}
We replay held-out recorded trajectories for GR00T on LIBERO (40 tasks,
376 episodes, $H=16$), VLA-JEPA on LIBERO (40 tasks, 581 episodes, $H=7$),
and Cosmos-Policy on RoboCasa (24 tasks, 376 episodes, $H=16$).
Task-specific CDFs are calibrated on the original validation episodes.
All signals use percentile threshold 0.80, gap factor 1.20, and requested
local fractions 0.2--0.9 with the same credit-based scheduler below.
The gap factor limits the interval from an advanced VLA call to the next one to 1.20 times
the nominal fixed-call interval, rounded up to a whole number of decisions.
We measure action RMSE in training-standardized coordinates. Local decisions
incur their prediction error against the recorded VLA chunk, whereas VLA
decisions have zero error against that reference. For each episode, we compare
each candidate with fixed scheduling up to the last boundary at which their
cumulative VLA-call counts agree, including the initial VLA call. We compute
RMSE from all retained chunks within each task, then average equally over
tasks and the eight fractions.

\begin{table}[htbp]
\centering
\caption{\textbf{Offline action-RMSE reduction relative to fixed scheduling (\%).}
Positive values indicate lower error on matched-call prefixes.
Tasks and requested local fractions are equally weighted.}
\label{tab:adaptive-offline}
\begin{tabular}{lrrr}
\toprule
Model / benchmark & Image change & Action change & Embedding risk \\
\midrule
GR00T / LIBERO & -0.22 & +0.52 & \textbf{+1.39} \\
VLA-JEPA / LIBERO & +0.52 & +1.78 & \textbf{+2.13} \\
Cosmos-Policy / RoboCasa & +0.22 & +1.33 & \textbf{+1.61} \\
\bottomrule
\end{tabular}
\end{table}

Table~\ref{tab:adaptive-offline} shows that embedding risk gives the largest
RMSE reduction in all three pairs. These offline results support choosing
risk-aware scheduling for the closed-loop comparison below. Offline replay
holds observations and executed histories fixed, so we use it to screen
routing signals rather than to infer task-success gains.

\paragraph{Risk score.}
We want to prioritize the VLA when the current state is unfamiliar to ULAP
and its local prediction may be less reliable. We use distance from training
representations as a proxy for this unfamiliarity. Let $\hat z_t$ be ULAP's
normalized current embedding and $\mathcal N_{32}(\hat z_t)$ its 32 nearest
same-task training embeddings. Their mean squared distance is
\begin{equation}
d_t=\frac{1}{32}\sum_{i\in\mathcal N_{32}(\hat z_t)}
\max\!\left(0,\,2-2\hat z_t^{\top}\hat z_i\right),
\qquad \hat z=\frac{z}{\lVert z\rVert_2}.
\end{equation}
To make distances comparable within each task, a frozen empirical CDF
$F_\tau$ converts $d_t$ to its percentile among that task's validation
distances, $r_t=F_\tau(d_t)$. A score of at least 0.80 flags a distance in
the highest 20\% of the calibration distribution and requests an earlier
VLA call. This is a risk score, not a failure probability.

\paragraph{Credit-based scheduling.}
The scheduler moves VLA calls toward high-risk decisions while tracking
the requested call budget. Think of credit as an allowance that accumulates
at each decision and is spent on VLA calls. A risky state can justify spending
that allowance early, but the resulting debt must not postpone the next VLA
call for too long. The initial chunk always uses the VLA and is outside this budget.

For requested local fraction $\rho=p/q$ in lowest terms, set $a=q-p$ and
initial credit $B_0=a-1$. Each subsequent decision adds $a$ credits,
$A_t=B_{t-1}+a$, and a VLA call spends $q$. Thus, credit accrues at a rate
corresponding to the target VLA fraction $a/q=1-\rho$.
If $A_t\geq q$, the scheduler makes a regular VLA call. Otherwise, it may
borrow credit to call the VLA early when $E_t$ holds below.
The gap factor $\gamma=1.20$ caps the post-borrowing interval at $G$
decisions, or 1.20 times the nominal VLA-call interval, rounded up:
\begin{equation}
G=\left\lceil\frac{\gamma}{1-\rho}\right\rceil,\qquad
E_t=(r_t\geq0.80)\land(A_t>0)\land(A_t-q\geq q-Ga).
\end{equation}
\begin{equation}
V_t=\begin{cases}
1, & A_t\geq q\text{ or }E_t,\\
0, & \text{otherwise},
\end{cases}
\qquad B_t=A_t-qV_t.
\end{equation}
The three conditions in $E_t$ require high risk, a positive credit balance
before borrowing, and enough remaining credit to fund another VLA call
within $G$ decisions. The last condition follows because the balance after
the early call is $A_t-q$, and the next $G$ decisions add $Ga$ credits.
Here $V_t=1$ selects the VLA, otherwise ULAP; $B_t$ is the balance carried
to the next decision. Disabling borrowing recovers fixed scheduling.
ULAP computes the risk embedding at every
post-initial boundary, and only the selected chunk is executed. Both
schedulers use the same predictor and matched evaluation episodes, with
16-step chunks for GR00T and Cosmos-Policy and seven-step chunks for VLA-JEPA.

\paragraph{Closed-loop results.}
We next test whether the offline error reduction translates into higher
task success when actions affect subsequent observations.
Figure~\ref{fig:ulap-cads-zoom} compares eight requested local fractions per
pair, with threshold 0.80 throughout. The horizontal axis is
$100\bigl(1-\sum_i C_i^{\mathrm{method}}/\sum_i C_i^{\mathrm{VLA}}\bigr)$,
where $C_i$ counts VLA calls in evaluation episode $i$, including failures.
Risk awareness improves GR00T success at seven fractions, but the
pattern is not consistent. On VLA-JEPA it improves four, ties one, and lowers
three; at fraction 0.9, success drops from 91.60\% to 90.30\% while call
reduction also drops from 86.52\% to 85.97\%. On RoboCasa, success and call
reduction both drop at fraction 0.5 (64.08\% to 63.25\% and 48.79\% to
46.85\%). We find no clear, consistent gain in the success--call-reduction
trade-off despite the favorable offline ranking. We therefore use fixed
scheduling as the standard configuration.

\begin{figure}[H]
\centering
\includegraphics[width=\linewidth]{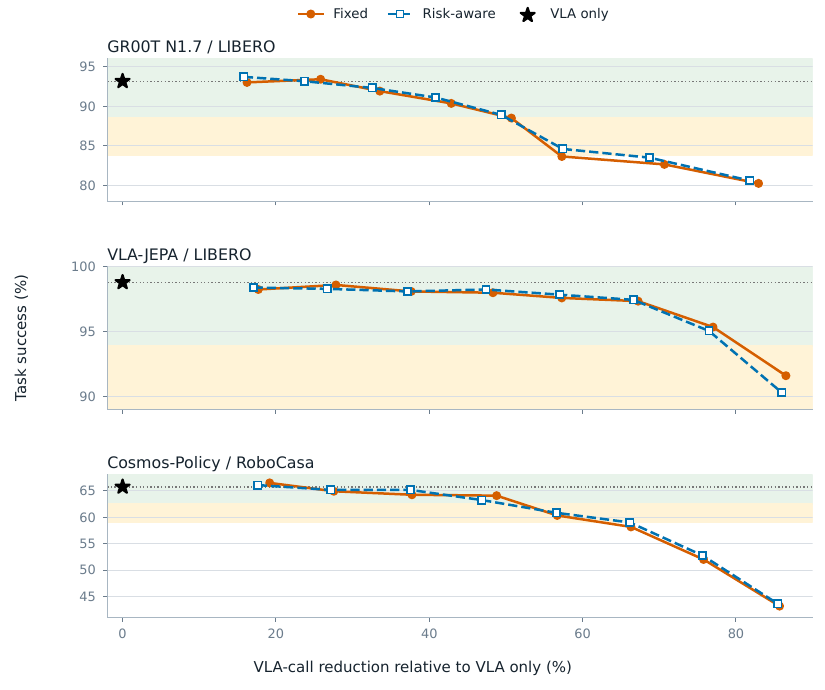}
\caption{\textbf{Closed-loop evaluation of the selected risk-aware scheduler.}
Success rate versus realized VLA-call reduction at requested local fractions
0.2--0.9. Black stars mark VLA-only baselines; green and amber bands indicate
at least 95\% and 90--95\% of the baseline success rate, respectively. Each LIBERO point
uses 2,000 episodes and each RoboCasa point 1,200; call totals include failures.}
\label{fig:ulap-cads-zoom}
\end{figure}

\clearpage
\section{Local prediction versus longer action-chunk execution}
\label{app:ulap-chunk-length}

Can simply executing longer chunks replace a local predictor? We test this
with SmolVLA~\citep{shukor2025smolvla}, whose evaluated checkpoint generates
50 actions per call. Several checkpoints in our main experiments instead
use short output horizons fixed during pretraining or benchmark fine-tuning,
so extending their execution horizon would require changing the base policy.
SmolVLA lets us isolate longer execution without retraining the VLA.
We increase the executed prefix from $H=20$ to 30, 40, or 50 steps,
whereas VLA-ULAP keeps $H=20$ and predicts fresh chunks at local decisions.

\begin{figure}[H]
\centering
\includegraphics{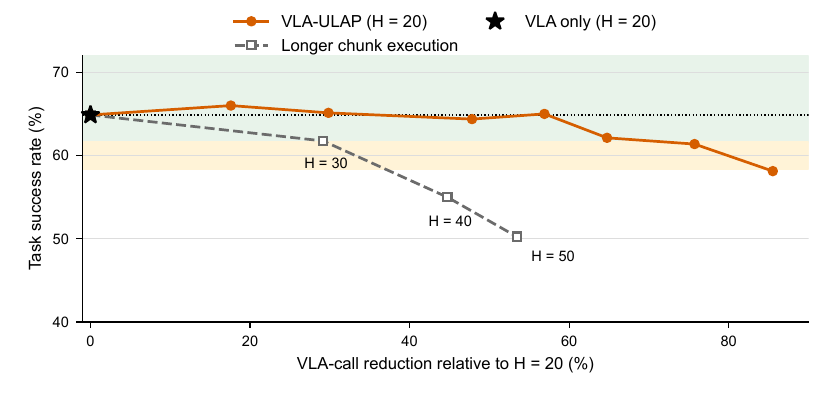}
\caption{\textbf{Local prediction retains success better than extending open-loop execution.}
SmolVLA on LIBERO; $H$ is the number of actions executed before a new
decision. Call reduction is relative to SmolVLA-only at $H=20$.
Green and amber bands denote $\geq$95\% and 90--95\% retention of that
baseline's success rate.}
\label{fig:ulap-chunk-length}
\end{figure}

\paragraph{Higher success at comparable or greater call reduction.}
Figure~\ref{fig:ulap-chunk-length} shows that longer open-loop execution
rapidly loses success. Increasing $H$ from 20 to 50 reduces calls by
53.4\%, but success falls from 64.88\% (519/800) to 50.25\% (402/800).
VLA-ULAP instead achieves 65.00\% success (520/800) at a larger
56.9\% call reduction: a \textbf{14.75~pp advantage} over
$H=50$. The same pattern holds at a more moderate reduction:
VLA-ULAP reaches 64.38\% (515/800) at 47.8\% reduction, versus
55.00\% (440/800) at 44.7\% reduction with $H=40$.
Even at 85.5\% reduction, VLA-ULAP retains 58.13\% success (465/800),
exceeding $H=50$ while using substantially fewer VLA calls.
These results support reducing calls through fresh, state-conditioned
predictions rather than simply postponing the next observation-conditioned decision.

\paragraph{Training and evaluation settings.}
Each condition uses the same 800 episodes: four LIBERO suites, ten tasks
per suite, and official initial states 0--19, with environment seed 1000.
The frozen \texttt{smolvla\_libero} checkpoint (revision \texttt{6721902b})
always generates 50 actions. ULAP uses current camera views, proprioception,
and the preceding 20 executed actions; one predictor per suite is trained on
official demonstrations with a 70/15/15 episode split and selected by
validation action RMSE. Training uses AdamW for 120 epochs, learning rate
$6\times10^{-4}$, and seed 0. Fixed scheduling evaluates local fractions
0.2, $1/3$, 0.5, 0.6, 0.7, 0.8, and 0.9 after an initial VLA call.
Success and call reduction use all evaluation episodes; reduction is
$1-C/C_{20}$, where $C$ and $C_{20}$ are total VLA calls in the evaluated
condition and the $H=20$ baseline, respectively.

\endgroup

\end{document}